\pdfoutput=1

\documentclass[11pt]{article}

\usepackage[]{acl}

\usepackage{times}
\usepackage{latexsym}

\usepackage[T1]{fontenc}

\usepackage[utf8]{inputenc}

\usepackage{microtype}

\usepackage{inconsolata}

\usepackage{graphicx}

\usepackage{appendix}

\usepackage{multirow}
\usepackage{amsmath}
\usepackage{caption}
\usepackage{subcaption}
\usepackage{booktabs}
\usepackage{tabularx}
\usepackage{colortbl}
\usepackage{lipsum}
\usepackage{amssymb}
\usepackage{amsmath}

\usepackage{algorithm}
\usepackage{algpseudocode}
\usepackage{subcaption}
\usepackage{array}

\newcolumntype{P}[1]{>{\raggedright\arraybackslash}p{#1}}

\usepackage{color,soul}
\usepackage{xcolor}
\usepackage[colorinlistoftodos,prependcaption,textsize=tiny]{todonotes}

\definecolor{mygreen}{RGB}{0, 128, 0}

\newcommand{\eat}[1]{}

\newcommand{\nursegpt}{\texttt{NurseLLM}}

\title{\includegraphics[scale=0.03]{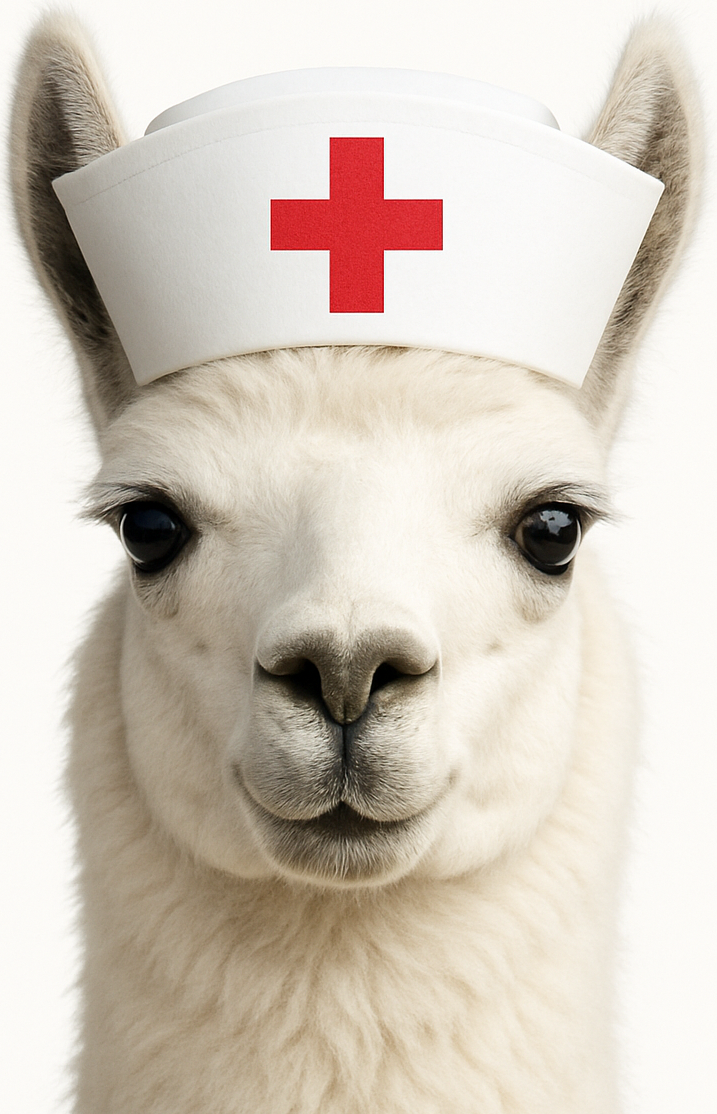} \nursegpt:\\The First Specialized Language Model for Nursing}

\author{Md Tawkat Islam Khondaker$^{\spadesuit}$$^{\heartsuit}$~~ Julia Harrington$^{\spadesuit}$$^{\diamondsuit}$~~ Shady Shehata$^{\spadesuit}$$^{\triangle}$
\\\\ 
\normalsize $^{\spadesuit}$Yourika Labs,
$^{\heartsuit}$The University of British Columbia,$^{\diamondsuit}$Western University,$^{\triangle}$University of Waterloo\\
\\
\texttt{\{mtawkat,jharrington,sshehata\}@yourika.ai}
}

\begin{document}
\maketitle
\begin{abstract}
Recent advancements in large language models (LLMs) have significantly transformed medical systems. However, their potential within specialized domains such as nursing remains largely underexplored. In this work, we introduce \nursegpt, the first nursing-specialized LLM tailored for multiple choice question-answering (MCQ) tasks. We develop a multi-stage data generation pipeline to build the first large scale nursing MCQ dataset to train LLMs on a broad spectrum of nursing topics. We further introduce multiple nursing benchmarks to enable rigorous evaluation. Our extensive experiments demonstrate that \nursegpt~outperforms SoTA general-purpose and medical-specialized LLMs of comparable size on different benchmarks, underscoring the importance of a specialized LLM for the nursing domain. Finally, we explore the role of reasoning and multi-agent collaboration systems in nursing, highlighting their promise for future research and applications.
\end{abstract}

\begin{figure*}[!t]
  \centering
  \includegraphics[width=0.9\textwidth]{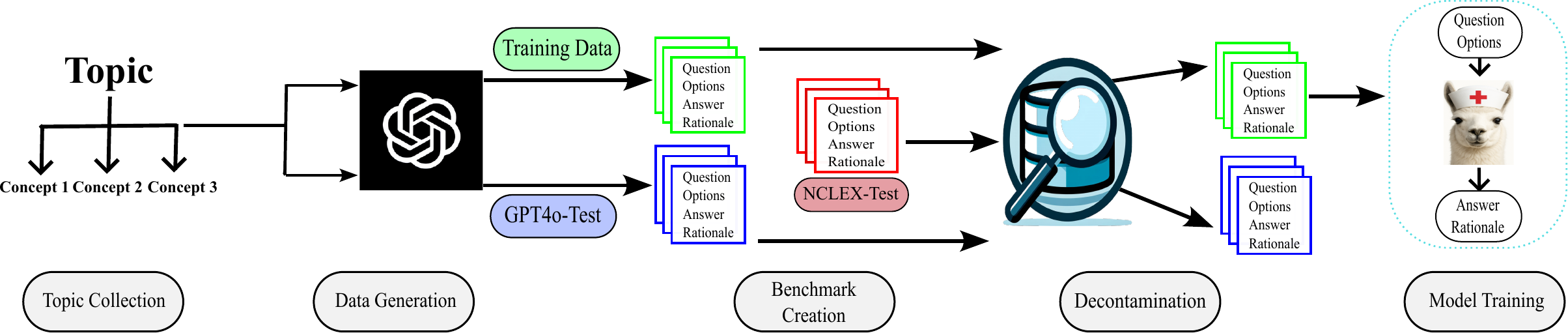}
  \caption{\label{fig:methodology}
  Overall methodology of \nursegpt~data collection and training. 
  }
\end{figure*}

\section{Introduction}
\label{sec:introduction}
The rapid progress of large language models (LLMs)~\cite{alpaca,vicuna,jiang2023mistral,qwen2.5,gemma2024,abdin2024phi,Dubey2024llama} has revolutionized general medical AI systems in clinical applications~\cite{med42v2,o1-replication,openbiollm}. However, the potential of LLMs in a specialized domain like nursing is yet to be explored.

\noindent Nursing and medicine, while essential for healthcare, operate within different domains of knowledge. Medicine usually adheres to the biomedical model, emphasizing diagnosis and treatment~\cite{crna_scope_rn_2022}, while nursing focuses on holistic, person-centered care that includes advocacy, education, and emotional support~\cite{crna_scope_rn_2022,Mazzotta2016BiomedicalAT}. The same patient may be viewed differently by nurses and physicians due to their unique approach to care~\cite{vreugdenhil2023}. With more than 28 million nurses worldwide, representing the largest segment of the global health workforce~\cite{WHO_2025_nursing_workforce}, the profession demands artificial intelligence (AI) tools designed specifically around its values and decision-making processes. A nursing-specific LLM must reflect this distinct perspective, moving beyond generic medical applications to truly support the holistic nature of nursing practice.

\noindent Another challenge of employing LLMs in nursing applications is the rigorous evaluation of the models on nursing questions. While the existing benchmarks~\cite{singhal2023large,medqa,medmcqa} evaluate LLMs on the generic medical knowledge, how these LLMs would situate themselves in nursing specialized question-answering exams like the NCLEX (National Council Licensure Examination) is still unexplored. The NCLEX\footnote{\href{https://cno.org/become-a-nurse/examinations/registered-nurse-examinations/about-nclex-rn}{College of Nurse of Ontario}} is a standardized entry-to-practice examination required for nurses in the United States and Canada to become licensed to practice as Registered Nurses (RN) or Licensed Practical Nurses (LPN). The NCLEX uses computer-adaptive testing (CAT) technology to administer the exam and measures clinical judgment and decision-making abilities through a variety of questions\footnote{\href{https://www.nclex.com/About.page}{National Council of State Boards of Nursing}}.

\noindent To address these gaps, we introduce \nursegpt~the first nursing specialized LLM to the best of our knowledge. In the process of developing \nursegpt, we create a large-scale NCLEX-equivalent nursing multiple-choice question-answering (QA) dataset comprising $125$K samples through a multi-stage data generation pipeline. We further develop three different nursing benchmarks for systematic evaluation of LLMs on nursing QA.

\noindent Our contributions can be summarized as follows:
\begin{enumerate}
    \item We develop \nursegpt, the first LLM specifically designed for nursing question-answering tasks. Given a nursing-related multiple-choice question, \nursegpt~provides not only the answer, but also a rationale to ensure the model's transparency and credibility.

    \item We design a multi-stage data generation pipeline to create the first large-scale and topic-diverse nursing question-answering dataset.

    \item For meticulous evaluation of LLMs on nursing QA, we further develop three different nursing MCQ benchmarks.

    \item Our extensive evaluation reveals that \nursegpt~outperforms SoTA generalized and medical expert LLMs of similar parameter magnitude on nursing benchmarks, highlighting the importance of a specialized LLM for the nursing domain.

    \item We further investigate the potential of reasoning and multi-agent collaboration systems within the nursing domain, revealing their promising implications for future advancements in the field.
\end{enumerate}

\section{Related Works}
\label{sec:related_works}

Currently, there are no LLMs developed specifically for nursing; most nursing applications rely on generic or medical LLMs. While LLMs have shown promise in optimizing nursing workflows, personalizing care, and improving healthcare accessibility~\cite{zhou_chatgpt_nursing}, existing research highlights significant limitations. Studies report that general LLMs, including \texttt{ChatGPT-3.5}, demonstrate indecisiveness in clinical assessments, suggest unnecessary diagnostic tests, and misunderstand nursing concepts, leading to a tendency for over-triage compared to human clinicians~\cite{saban2024chatgptclinical}. Moreover, these models depend heavily on pre-existing data, which may become outdated without continual updates, compromising their reliability and potentially propagating obsolete or inaccurate information~\cite{Huang2023ChatGPTNursing,zhou_chatgpt_nursing}. Additionally, the lack of transparency in training data and model processes raises concerns regarding accuracy, safety, and usability in clinical practice~\cite{Glauberman2023AIinNursing,gonzalez2024chatgptnursingeducation}. Issues such as implicit bias and data-shift can further exacerbate inequities in care, emphasizing the urgent need for nursing-specific LLMs optimized through rigorous research to support clinical decision-making effectively~\cite{zhou_chatgpt_nursing,saban2024chatgptclinical}.\\
\noindent \textbf{LLMs in Medical Application.} In recent years, researchers have developed a diverse range of LLMs~\cite{med42v2,openbiollm,medalpaca} focusing on medical and clinical settings. Examples of such LLMs include BioMistral~\cite{labrak2024biomistral}, BioMedGPT~\cite{biomedgpt}, Meditron~\cite{meditron}. A number of medical LLMs~\cite{aloe,Yu2025FineMedLMo1ET} have also been created on the top of a generalized LLM like LLaMA~\cite{Dubey2024llama}. More recently, inspired by the strong performance of the reasoning LLMs, especially in medical applications~\cite {o1-replication},~\citet{chen2024huatuogpto1} develop HuatuoGPT-o1, providing valuable insights for enhancing the complex medical reasoning capabilities of LLMs.

\noindent LLMs are commonly benchmarked on the existing MCQ datasets on medical knowledge like MedQA~\cite{medqa}, MedMCQA~\cite{medmcqa}, and MMLU-Medical~\cite{hendryckstest2021mmlu}. Recently,~\citet{zuo2025medxpertqa} introduce more challenging benchmark MedXpertQA to evaluate expert-level and advanced reasoning capabilities of LLMs spanning across $11$ body systems. However, recent studies~\cite{Chen2024ClinicalBenchCL,liu-etal-2024-large,jeong-etal-2024-medical} expose a critical gap of medical LLMs on the clinical domain.

\noindent Despite recent advancements in medical LLMs, existing research has largely overlooked the development and evaluation of LLMs specifically tailored to the nursing domain. This gap in the literature motivates us to develop \nursegpt~and introduce novel benchmarks specifically curated to address the unique needs and challenges within nursing practice.

\section{Methodology}
\label{sec:methodology}
We present our overall methodology of NCLEX data generation and \nursegpt~training in Figure~\ref{fig:methodology}. Now, we describe each component below:

\begin{figure}[!ht]
    \centering
  \includegraphics[width=0.4\columnwidth]{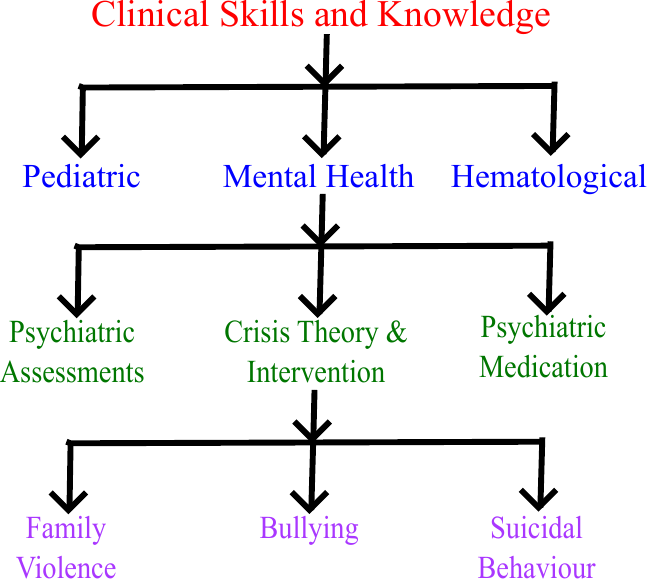}
  \caption{\label{fig:taxonomy}
  Example of a sample taxonomy (\textcolor{red}{\textit{Specialization}} $\rightarrow$ \textcolor{blue}{\textit{Domain}} $\rightarrow$ \textcolor{mygreen}{\textit{Topic}} $\rightarrow$ \textcolor{violet}{\textit{Concept}}) for the specialization of \textit{Clinical Skills and Knowledge}.
  }
\end{figure}

\subsection{Topic Collection}

In order to generate diverse NCLEX question-answer pairs, we first produced a comprehensive and exhaustive taxonomy of NCLEX topics. Two 
Registered Nurses with NCLEX expertise defined the high-level classification of \textit{specializations} based on the NCLEX questions available in books and web corpora. Each specialization is further categorized into \textit{domains} based on the expertise required for a nurse to be classified into that respective specialization. Then, analyzing the questions, they created the \textit{topic} list for each domain and diversified the topics into different \textit{concepts}. As a result, we created a comprehensive and diverse list of $7$ \textit{Specializations}, $60$ \textit{Domains}, $232$ \textit{Topics}, and $1,830$ \textit{Concepts}. A sample of this taxonomy is presented in Figure~\ref{fig:taxonomy}. We share the full taxonomy in Appendix~\ref{appendix:topic_list}.

\subsection{Synthetic NCLEX QA Generation}
\label{sec:synthetic_nclex_qa_generation}

\begin{figure}[!ht]
    \centering
  \includegraphics[width=0.7\columnwidth]{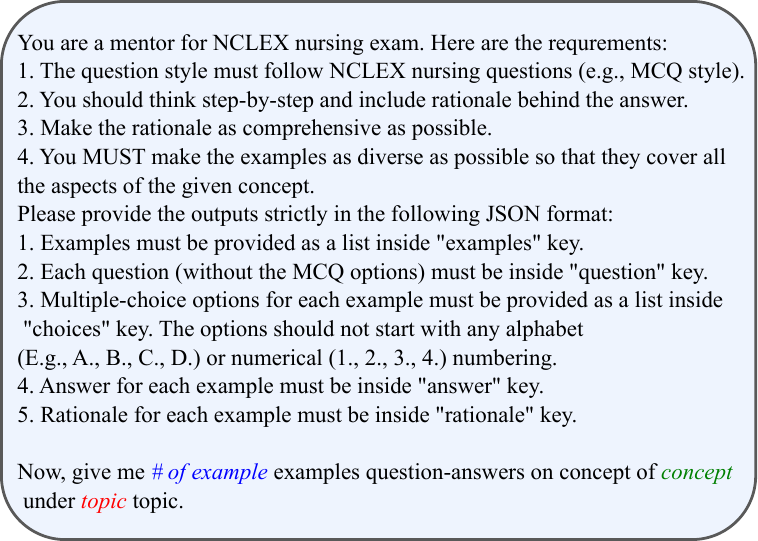}
  \caption{\label{fig:nclex_prompt}
  Prompt template used for synthetic data generation.
  }
\end{figure}

After creating the diverse list of concepts, we prompt GPT-4o to generate corresponding NCLEX MCQ question-answer (QA) from those concepts. The prompt to generate QA is given in Figure~\ref{fig:nclex_prompt}. Similar to the real NCLEX exam, we also generate the rationale that justifies the correct answer. We use this rationale later to train \nursegpt~with chain-of-thought (CoT)~\cite{wei2022chain} reasoning. We provide the samples of synthetic NCLEX MCQ in Appendix~\ref{appendix:nclex_examples}.\\

\noindent \textbf{GPT4o-Test Benchmark Generation}\\
\noindent Along with generating the synthetic data for training, we also generate a synthetic benchmark for testing the models. For this purpose, we generate one NCLEX QA per concept with GPT-4o and use these $1,830$ samples as \textbf{GPT4o-Test} benchmark.

\subsection{Benchmarks Creation}

To rigorously test LLMs on the nursing tasks, we develop multiple NCLEX-based benchmarks. First, we develop a human-labeled gold NCLEX benchmark comprising $1,726$ NCLEX questions (\textbf{NCLEX-Test}). Our Registered Nurses manually create these $1,726$ NCLEX questions based on NCLEX expertise and knowledge gained from experience as educators. They further verified the correct answer as well as the rationales that justify the answer. We use this NCLEX-Test as our prime benchmark to test LLMs.

We further employ GPT-4o to extract nursing exclusive questions from MultiMedQA~\cite{singhal2023large}. MultiMedQA is a collection consisting of multiple generic medical QA datasets. We prompt GPT-4o to identify the QA samples exclusively relevant to nursing topics (prompt for this topic modeling is presented in Appendix~\ref{appendix:multinurseqa_prompt}). In this way, we mine a subset of MultiMedQA, namely \textbf{MultiNurseQA} to evaluate LLMs on the nursing-related questions.\\
\noindent Therefore, as summarized in Section~\ref{sec:evaluation_benchmarks}, we develop three distinct benchmark datasets: NCLEX-Test, GPT4o-Test, MultiNurseQA, and in conjunction with MultiMedQA, we provide a comprehensive and rigorous framework for evaluating LLMs across both nursing-specific and general medical domains.

\subsection{Decontamination}

On one hand, it is important to decontaminate the trainings samples from the test set to ensure the generalization of \nursegpt, on the other hand, the generated training samples need to be unique to ensure diversity. However, comparing each question against a large-scale dataset is computationally expensive. Hence, to make the decontamination process efficient, we devise a two-step filtration. \textbf{(i)} First we decontaminate each generated sample against the whole \textbf{GPT4o-Test} and \textbf{NCLEX-Test} benchmarks to make sure that there is no overlapping between the train and the test datasets. \textbf{(ii)} If a sample passes the first step, then we compare this sample against the pool of the same concept, meaning that we compare this sample against the already generated training samples assigned to the same concept. Since each concept is unique, samples from different concepts will likely also be diverse. Therefore, we can avoid comparing a sample against a pool of samples from different concepts. Following~\citet{alpaca}, we use the ROUGE-L score for the sample comparison. We apply the threshold of $0.5$ and filter out any sample that exceeds this score.

\section{Experimental Setup}
\label{sec:experintal_setup}
\subsection{Model Training}

For finetuning of \nursegpt, we use \texttt{Llama3-Med42-8B}~\cite{med42v2} as the base model. The maximum length of the model was set to $2,048$. The batch size is set to $1$ with gradient accumulation step at $8$. We train the model for $2$ epochs with cosine scheduler and a learning rate of $2e$-$5$. We use QLoRA~\cite{qlora} with $4$-bit quantization for the efficient utilization of GPU. The model is trained with an NVIDIA 10G GPU on AWS SageMaker for $4$ days.

\subsection{Model Merging}

Finetuning LLMs on new data can cause catastrophic forgetting~\cite{li-etal-2022-forget,scialom-etal-2022-forgetting}. To address this issue, we apply MergeKit to merge the finetuned LLM with the base Med42 using the DARE~\cite{dare_merging} method. We set the merging proportion to $60$\% upon inspecting the model's performance on the validation set. Merging with the base model allows \nursegpt~to preserve the generic medical capability while adapting to the new nursing knowledge.

\subsection{Evaluation Benchmarks}
\label{sec:evaluation_benchmarks}

To evaluate \nursegpt~rigorously on the NCLEX task, we use the following four benchmarks:

\begin{enumerate}
    \item \textbf{NCLEX-Test:} Human-labeled datasets comprising of $1,726$ MCQ questions.
    \item \textbf{GPT4o-Test:} GPT4o-generated $1,830$ NCLEX MCQ questions.
    \item \textbf{MultiMedQA:} A comprehensive suit~\cite{singhal2023large} consisting multiple medical question-answering datasets.
    \item \textbf{MultiNurseQA:} A subset of MultiMedQA topic modeled exclusively for nursing questions.
\end{enumerate}

\subsection{Baselines}
\label{sec:baselines}

We compare \nursegpt~against the following baselines consisting of both generic and medical specialized LLMs:\\
\textbf{(1)} \textbf{Med42} (Llama3-Med42-8B)~\cite{med42v2}; \textbf{(2)} \textbf{Gemma} (google/gemma-7b)~\cite{gemma2024}; \textbf{(3)} \textbf{Apollo} (FreedomIntelligence/Apollo-7B)~\cite{wang2024apollo}; \textbf{(4)} \textbf{LLaMA} (meta-llama/Meta-Llama-3-8B-Instruct)~\cite{Dubey2024llama}; \textbf{(5)} \textbf{OpenBioLLM} (aaditya/Llama3-OpenBioLLM-8B)~\cite{openbiollm}; \textbf{(6)} \textbf{Mistral} (mistralai/Mistral-7B-Instruct-v0.1)~\cite{jiang2023mistral}; \textbf{(7)} \textbf{BioMistral} (BioMistral/BioMistral-7B)~\cite{labrak2024biomistral}; \textbf{(8)} \textbf{MedAlpaca} (medalpaca/medalpaca-7b)~\cite{medalpaca}; \textbf{(9)} \textbf{DeepSeek-LLaMA} (deepseek-ai/DeepSeek-R1-Distill-Llama-8B)~\cite{deepseek-r1}.\\

\noindent All the LLMs are hosted locally for conducting the experiments. Please note that we do not include any closed-source LLMs like ChatGPT in our experiments because (i) The size of the closed-source models is not comparable to our baseline, and (ii) We do not intend to pass our NCLEX-Test benchmark to closed-source LLMs due to proprietary reasons.

\section{Results}
\label{sec:results}
 \subsection{Performance on NCLEX-Test}

\begin{figure}[t]
    \centering
  \includegraphics[width=0.8\columnwidth]{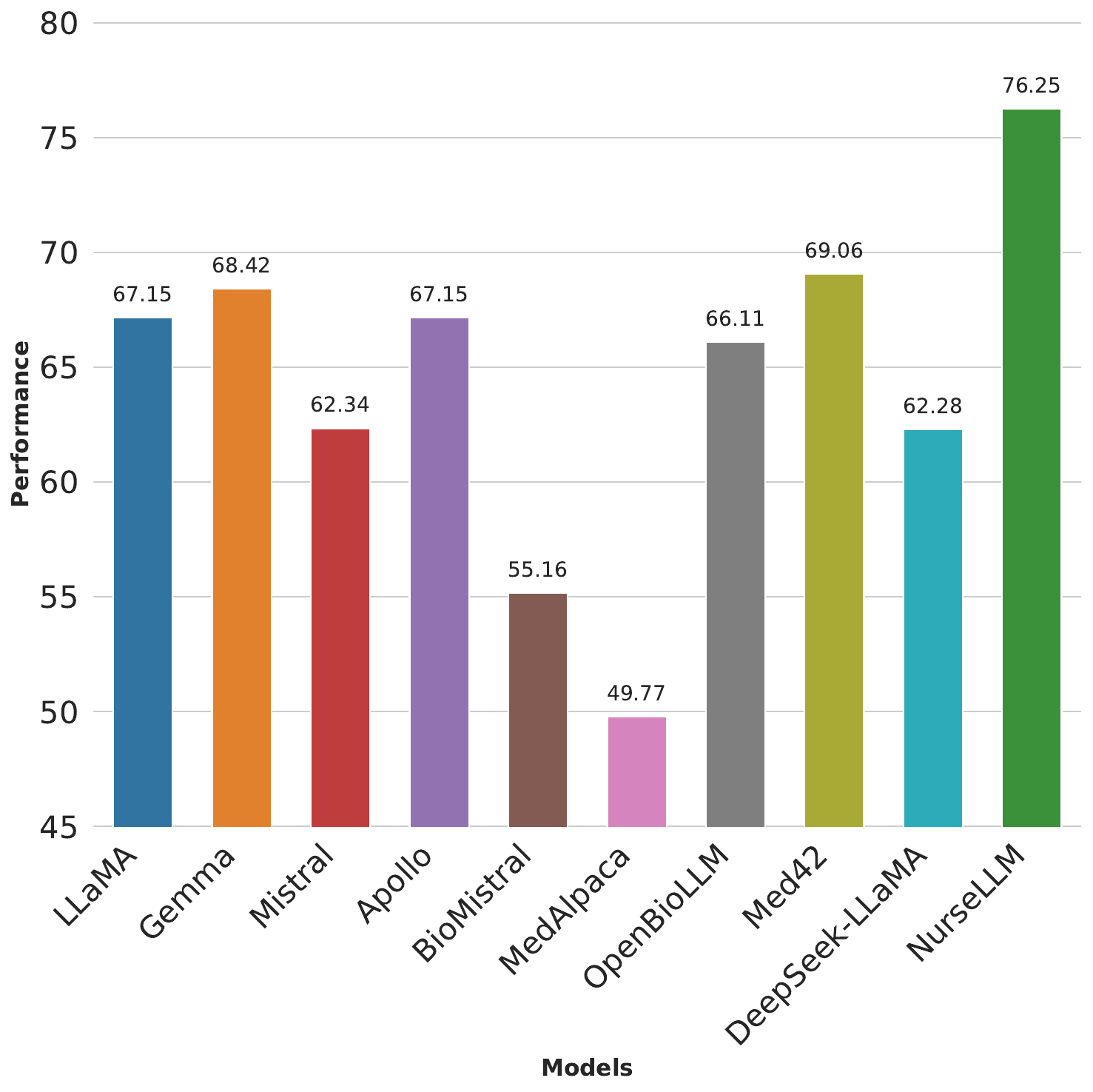}
  \caption{\label{fig:performance_nclex}
  Performance of the models on the NCLEX-Test Benchmark.
  }
\end{figure}

We present the performance of the LLMs on the NCLEX-Test benchmark in Figure~\ref{fig:performance_nclex}. As evident, \nursegpt~outperforms all the open-source LLMs with compatible parameters. Noticeably, even LLMs specialized in the medical domain (E.g., Med42, OpenBioLLM) cannot reach the accuracy of $70$\%, whereas \nursegpt~scores above $76$\%. This finding highlights \textbf{(1)} The importance of specialized LLMs for nursing domain, \textbf{(2)} The efficacy of our synthetic diverse dataset that helps \nursegpt~to improve the performance, and \textbf{(3)} The necessity of challenging benchmark like NCLEX-Test for the rigorous evaluation of medical LLMs on nursing domain.

\subsection{Performance on GPT4o-Test}

\begin{figure}[!t]
    \centering
  \includegraphics[width=0.8\columnwidth]{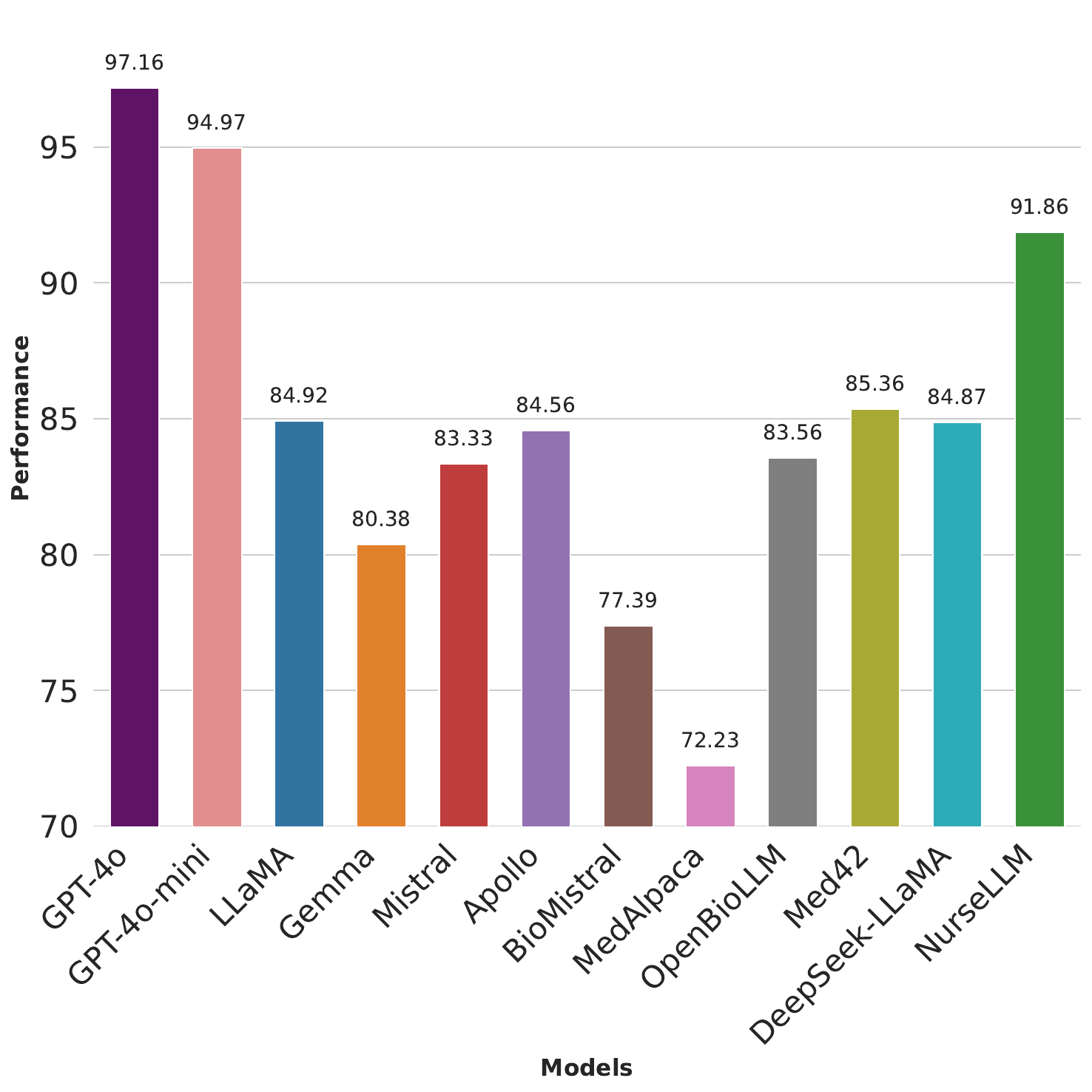}
  \caption{\label{fig:performance_gpt4o}
  Performance of the models on GPT4o-Test Benchmark.
  }
\end{figure}

We show the performance of the models on GPT-4o created NCLEX benchmark in Figure~\ref{fig:performance_gpt4o}. Although the benchmark is GPT-4o generated, GPT-4o cannot reach the perfect accuracy due to the stochastic nature of the model. As evident, \nursegpt~with $8$B parameters achieves over $91$\% accuracy close to the performance of GPT-4o-mini ($94.97$\%). However, other baseline LLMs struggle to compete with \nursegpt, which again signifies the necessity for a nursing specialized LLM.

\subsection{Performance on MultiMedQA}

\begin{table*}[!t]
\centering
\resizebox{\linewidth}{!}{
\begin{tabular}{lcccccccccc}
\hline
\textbf{Models} & \textbf{Avg.} & \textbf{Clinical KG} & \textbf{Medical Genetics} & \textbf{Anatomy} & \textbf{Pro Medicine} & \textbf{College Biology} & \textbf{College Medicine} & \textbf{MedQA 4 opts} & \textbf{PubMedQA} & \textbf{MedMCQA} \\
\hline
LLaMA & 69.19 & 68.30 & 83.00 & 68.89 & 70.22 & 78.47 & 61.85 & 59.70 & 74.80 & 57.52 \\
Gemma & 64.18 & 69.81 & 70.00 & 59.26 & 66.18 & 79.86 & 60.12 & 47.21 & 76.20 & 48.96 \\
Mistral & 62.85 & 68.68 & 71.00 & 55.56 & 68.38 & 68.06 & 59.54 & 50.82 & 75.40 & 48.20 \\
Apollo & 60.00 & 62.26 & 72.00 & 61.48 & 69.12 & 70.83 & 55.49 & 55.22 & 39.80 & 53.77 \\
BioMistral & 58.97 & 59.90 & 64.00 & 56.50 & 60.40 & 59.00 & 54.70 & 50.60 & 77.50 & 48.10 \\
MedAlpaca & 58.04 & 57.36 & 69.00 & 57.04 & 67.28 & 65.28 & 54.34 & 41.71 & 72.80 & 37.51 \\
OpenBioLLM & 71.99 & \textbf{78.49} & \textbf{84.00} & 68.89 & \textbf{78.68} & \textbf{83.33} & 67.05 & 58.05 & 71.80 & 57.59 \\
Med42 & \textbf{72.41} & 76.23 & 78.00 & \textbf{71.85} & 76.84 & 80.56 & \textbf{68.21} & 62.77 & 77.00 & 60.24 \\
DeepSeek-LLaMA & 65.45 & 67.23 & 76.00 & 59.33 & 67.85 & 70.22 & 58.14 & 55.79 & 77.40 & 57.23 \\
\nursegpt & 71.01 & 76.98 & 80.00 & 68.89 & 73.53 & 76.39 & 63.01 & \textbf{62.84} & \textbf{77.80} & \textbf{59.65} \\
\hline
\end{tabular}
}
\caption{
\label{table:performance_multimedqa}
Performance of the models on the MultiMedQA benchmark.
}
\end{table*}

We compare the models' performance on the generic medical benchmark MultiMedQA in Table~\ref{table:performance_multimedqa}. As we can see, in overall performance, medical specialized models like Med42 dominates \nursegpt~on this benchmark. This is unsurprising since medical specialized models are specifically pre- and post-trained on medical data. Although \nursegpt~utilizes Med42 as the base model, the model unlearns some of its medical knowledge while post-training on nursing-specific data because of the catastrophic forgetting~\cite{li-etal-2022-forget} which happens during the domain adaptation. However, \nursegpt~is still able to maintain comparable performance with other baselines. This indicates that, in addition to functioning as a nursing expert, \nursegpt~is also capable of providing answers on generic medical knowledge.

\subsection{Performance on MultiNurseQA}

\begin{table*}[t]
\centering
\resizebox{0.7\linewidth}{!}{
\begin{tabular}{lcccccccccc}
\hline
\textbf{Models} & \textbf{Avg.} & \textbf{Clinical KG} & \textbf{Anatomy} & \textbf{Pro Medicine} & \textbf{MedQA 4 opts} & \textbf{PubMedQA} & \textbf{MedMCQA} \\
\hline
LLaMA                                            & 65.05                             & 76.58                                    & 40                                   & 70.73                                     & 67.41                                     & 78.05                                 & 57.54                                \\
Gemma                                            & 58.32                             & 66.06                                    & 40                                   & 57.14                                     & 68.29                                     & 73.17                                 & 45.26                                \\
Mistral                                          & 53.52                             & 69.72                                    & 20                                   & 42.86                                     & 63.41                                     & 73.17                                 & 51.93                                \\
Apollo                                           & 55.88                             & 63.56                                    & 20                                   & 61.47                                     & 63.58                                     & 70.18                                 & 56.48                                \\
BioMistral                                       & 58.66                             & 60.53                                    & 40                                   & 56.79                                     & 68.24                                     & 75.56                                 & 50.81                                \\
MedAlpaca                                        & 54.42                             & 58.23                                    & 40                                   & 54.52                                     & 63.28                                     & 71.93                                 & 38.58                                \\
OpenBioLLM                                       & 69.23                             & 76.15                                    & \textbf{60}                          & 71.84                                     & 68.29                                     & 78.05                                 & 61.05                                \\
Med42                                            & \textbf{69.96}                    & 77.06                                    & \textbf{60}                          & 71.43                                     & \textbf{75.61}                            & 73.17                                 & \textbf{62.46}                       \\
DeepSeek-LLaMA                                   & 66.86                             & 68.17                                    & \textbf{60}                          & 65.55                                     & 71.83                                     & 76.34                                 & 59.25                                \\
\nursegpt & 69.77                             & \textbf{78.90}                           & 40                                   & \textbf{85.71}                            & 73.17                                     & \textbf{80.49}                        & 60.35\\                               

\hline
\end{tabular}
}
\caption{
\label{table:performance_multinurseqa}
Performance of the models on MultiNurseQA benchmark.
}
\end{table*}

We further compare the models' performance on the `nursing' subset of MultiMedQA, namely MultiNurseQA in Table~\ref{table:performance_multinurseqa}. As observed, Med42 continues to outperform \nursegpt, consistent with MultiMedQA. However, \nursegpt~achieves the second best performance on this nursing benchmark, trailing Med42 by only a narrow margin ($0.19$\%). When compared with the results from MultiMedQA (Table~\ref{table:performance_multimedqa}), a general decline in model performance is evident on MultiNurseQA. Notably, \nursegpt~experiences the smallest performance degradation, while the other models suffer more substantial declines.  This difference in performance further reveals the models' relative deficiency in the nursing-specific domain. 

\section{Analyses}
\label{sec:analyses}

\subsection{Impact of Reasoning-Incorporated Training}

\begin{figure}[!ht]
    \centering
  \includegraphics[width=\columnwidth]{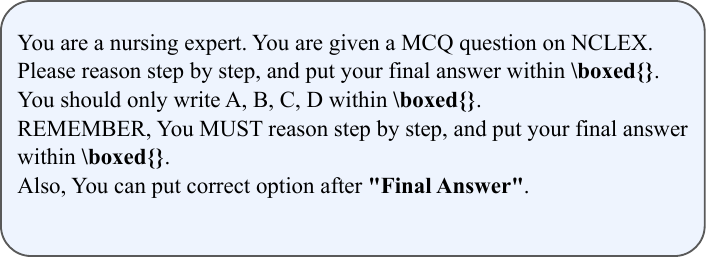}
  \caption{\label{fig:prompt_deepseek}
  Prompt to generate reasoning traces with \texttt{DeepSeek-R1-Distill-Qwen-32B} for NCLEX training data.
  }
\end{figure}

Large reasoning models~\cite{Zhong2024EvaluationOO, o1-replication, deepseek-r1} have shown remarkable performance on complex reasoning tasks in recent years. Inspired by this, we attempt to study how incorporating reasoning during the inference time~\cite{snell2025scaling}. First, we sample $5,000$ data from our synthetic post-training dataset. Then, we prompt (Figure~\ref{fig:prompt_deepseek}) \texttt{DeepSeek-R1-Distill-Qwen-32B} model to generate the reasoning traces. We filter out those samples where the \texttt{DeepSeek-R1-Distill-Qwen-32B} does not agree with GPT-4o on the final answer, resulting in $4,683$ final samples. After the filtration,  A Med42 model is then supervised finetuned on these MCQ samples with reasoning traces.

\noindent We evaluate the reasoning version of \nursegpt~on our NCLEX-Test benchmark. We observe that the reasoning model scores $72.01$\%, trailing SoTA \nursegpt~by $4.24$\% while using only $1$/$30$\textsuperscript{th} samples of \nursegpt. This finding highlights the promising potential of leveraging the reasoning capabilities of models. As a future direction, we are currently expanding the reasoning-incorporated nursing post-training dataset to develop a nursing specialized large reasoning model. An example of the response generated by \nursegpt~is presented in Appendix~\ref{appendix:reasoning_example}.

\subsection{\nursegpt~as Multi-Agent System}

\begin{figure}[!ht]
    \centering
  \includegraphics[width=\columnwidth]{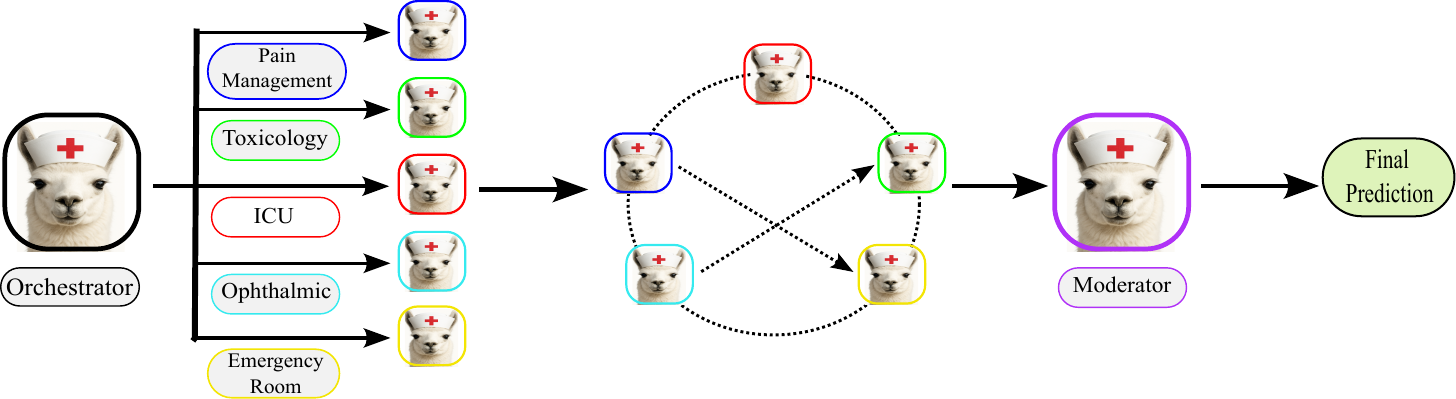}
  \caption{\label{fig:mdagent_nursegpt}
  Workflow of our multi-agent collaborative system.
  }
\end{figure}

The exploration of multi-agent systems (MAS) has gained increasing attraction among researchers~\cite{adas,joty-survey-llm-2025}. Inspired by~\citet{kim2024mdagents}, we deploy \nursegpt~as a multi-agent collaborative system (Figure~\ref{fig:mdagent_nursegpt}) to study its capability in the nursing domain. The workflow of our MAS is designed as follows: \textbf{(1)} upon receiving a nursing question an \textit{Orchestrator} LLM first determines the difficulty of the question. \textbf{(2)} If the difficulty is determined as easy, the question is then routed to a single LLM. Otherwise, the question is routed to a group of LLMs dynamically assigned by the \textit{Orchestrator}, where each LLM is an \textit{Expert} in different nursing specializations. The \textit{Experts} then involve in a multi-round collaborative debate. Finally, a \textit{Moderator} analyzes discussions among the \textit{Experts} and provides the final answer as the verdict. We present responses from MAS in Appendix~\ref{appendix:sample_mas}.

\noindent We evaluate the performance of MAS on NCLEX-Test benchmark. We find that MAS exhibits $79.87$\% accuracy improving over SoTA \nursegpt~by $3.62$\%. This finding demonstrates that we can further improve the capability of nursing LLMs by deploying in a multi-agent collaborative system. We provide runtime comparison across different model configurations in Appendix~\ref{appendix:runtime_comparison}.

\section{Ablation Study on Training Data Quantity}
\label{sec:ablation_data_quantity}

\begin{figure}[!ht]
    \centering
  \includegraphics[width=\columnwidth]{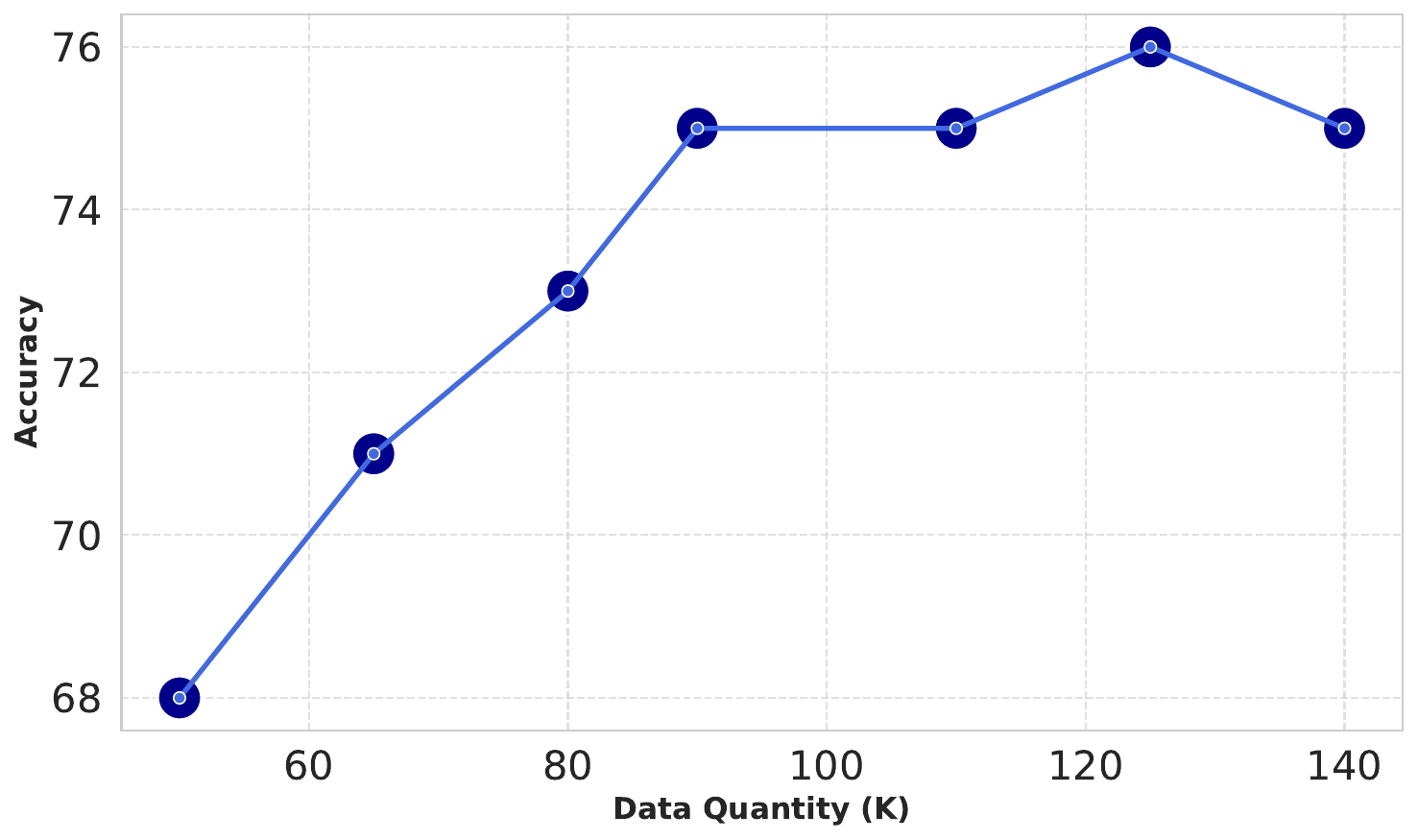}
  \caption{\label{fig:ablation_data_quantity}
  Ablation study on the performance of the model trained on different sample size.
  }
\end{figure}

We conduct an ablation study to investigate the relation between the quantity of post-training data and the performance of \nursegpt. For this purpose, we use a small human-labeled dataset of $100$ samples as our validation set. We present our result in Figure~\ref{fig:ablation_data_quantity}. As noticed, the model's performance improves monotonically with the increase amount of data. Nevertheless, the improve rate decreases after $90$K and at $125$K the performance reaches at a peak. Hence, considering the cost of high-quality synthetic data acquisition, we opted to limit the synthetic dataset to a sample size of $140$K.

\section{Ablation Study on Base Models}
\label{sec:ablation_base_model}

\begin{table}[ht]
\centering
\resizebox{0.7\columnwidth}{!}{
\begin{tabular}{lc}
\hline
\textbf{Base Models} & \textbf{Accuracy}\\
\hline
LLaMA                                            & 72\\
OpenBioLLM                                       & 73\\
Med42                                            & \textbf{76}\\           
\hline
\end{tabular}
}
\caption{
\label{table:ablation_base_models}
Ablation study on the base models used for finetuning.
}
\end{table}

We further study the impact of employing different LLMs as base models on the performance for the NCLEX benchmark. For this, we experiment with Med42, LLaMA-3.1, OpenBioLLM as the base models. We use the same validation set from Section~\ref{sec:ablation_data_quantity} for this study. We show the performance of \nursegpt~with different base models in Table~\ref{table:ablation_base_models}. As we can see, Med42 as a base model produces the best performance, followed by OpenBioLLM. This result reflects that a generic medical specialized LLM as a base model can lead to a better performance when adapting the model to specifically to the nursing domain.

\section{Conclusion}
\label{sec:conclusion}

In this work, we introduce \nursegpt, the first LLM specifically tailored for nursing question-answering. We design a multi-stage data generation pipeline to construct the first large-scale, diverse nursing QA dataset. We further develop three rigorous NCLEX-style benchmarks for the comprehensive evaluation of LLMS on the nursing domain. Our extensive experiments demonstrate that \nursegpt~significantly outperforms strong generalized and medical LLMs of comparable size, underscoring the necessity of domain specialization in the nursing field. Finally, we explore the potential of reasoning-driven and multi-agent collaborative approaches within the nursing domain, showcasing their potential for advancing intelligent systems in nursing applications.

\section{Limitations and Ethics Statement}
\subsection{Limitations}

\noindent \textbf{Data Generation Timeline.} In this work, we use \texttt{GPT-4o} version from May 2024. Since the model can be updated on a regular interval, the prompting strategy and the data generation pipeline discussed in Section~\ref{sec:methodology} should be treated accordingly, since the model's responses can change over time~\cite{chen2023chatgptchange}.

\noindent \textbf{Data Quality.} We develop an automated data generation pipeline to create a large scale diverse nursing-focused QA dataset. (Section~\ref{sec:methodology}). Although this automated pipeline makes the overall data generation process scalable, it comes at the risk of allowing low-quality data in the corpus. Hence, we suggest human inspection to remove any sort of potential vulnerability and maintain a standard quality of the corpus.

\noindent \textbf{Model Responses.} We show that \nursegpt~exhibits impressive performance on NCLEX questions compared to the other similar baselines. However, looking at the results (Section~\ref{sec:results}), we believe there is still room for improvement for the models in terms of accuracy. This also indicates, LLM can produce inaccurate responses while answering to the nursing questions. Given the sensitiveness of the field, we recommend that \nursegpt~should be couched with caution before deployment. 

\noindent \textbf{Model Evaluation.} We only use open-sourced models while conducting evaluation on our NCLEX benchmarks. As discussed in Section~\ref{sec:experintal_setup}, we omit evaluating closed-source LLMs  since (i) the size of the closed-source models is not comparable to our baselines, and (ii) we do not intend to expose the NCLEX-Test benchmark to closed-source LLMs due to the proprietary reasons. However, we aim to release our synthetic test benchmarks as well as the training dataset upon acceptance.

\subsection{Ethics Statement}
\noindent \textbf{Data Collection.} As mentioned in Section~\ref{sec:methodology}, we create a large-scale training dataset from a wide range of diverse nursing topics. We share the complete topic list in Appendix~\ref{appendix:topic_list} to facilitate the  NCLEX samples generation. 

\noindent \textbf{Intended Use of \nursegpt.} The application of \nursegpt~is to solve NCLEX-style nursing MCQ question while also providing corresponding rationale. We aim to help researchers build nursing-specialized LLMs. \nursegpt~can also be regarded as a promising baseline to develop more robust and effective LLMs suited for nursing applications.

\noindent \textbf{Potential Misuse and Bias.}
Our NCLEX corpus and \nursegpt~model can potentially be misused to generate inaccurate and biased clinical data. For these reasons, we recommend that \nursegpt~not be used in applications without careful prior consideration of potential misuse and bias.

\bibliography{custom}

\clearpage
\appendix
\appendixpage
\addappheadtotoc
\counterwithin{figure}{section}
\counterwithin{table}{section}

\section{Extracting Nursing Question from MultiMedQA}
\label{appendix:multinurseqa_prompt}

\begin{figure}[!ht]
    \centering
  \includegraphics[width=\columnwidth]{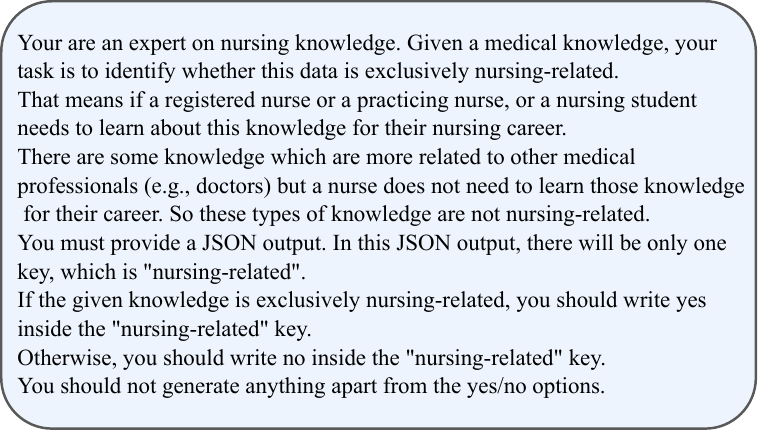}
  \caption{\label{fig:multinurseqa_prompt}
  Prompt used for extracting nursing-relevant QA samples from MultiMedQA.
  }
\end{figure}

\noindent We prompt GPT-4o to identify the question-answering samples exclusively relevant to nursing topics. We present the prompt for this topic modeling in Figure~\ref{fig:multinurseqa_prompt}. In this way, we mine a subset of MultiMedQA, namely \textbf{MultiNurseQA}, to evaluate LLMs on the nursing-specific questions.

\section{Runtime Comparison}
\label{appendix:runtime_comparison}

We provide the comparison of runtimes of different models in Table~\ref{tab:runtime}.

\begin{table}[h]
\centering
\begin{tabular}{lcc}
\toprule
\textbf{Model} & \textbf{Runtime (sec)} & \textbf{Comparison} \\
\midrule
Single-LLM   & 5.6  & 1$\times$ \\
Reasoning-LLM   & 8.8  & 1.57$\times$ \\
MAS-easy     & 6.1  & 1.09$\times$ \\
MAS-hard     & 28.3 & 5.05$\times$ \\
\bottomrule
\end{tabular}
\caption{Runtime comparison across different model configurations.}
\label{tab:runtime}
\end{table}

As evident, reasoning-based LLMs require approximately 1.5$\times$ more computational time due to the additional step of generating intermediate reasoning before producing the final prediction. For easy questions, our MAS framework incurs comparable latency to a single-LLM setup, since such questions are routed directly to a single LLM by the orchestrator. In this case, the only overhead arises from the orchestrator’s difficulty assessment. In contrast, for hard questions, MAS requires about 5$\times$ more time than a single-LLM baseline, as the architecture initiates a multi-round debate among multiple LLMs before converging on a final answer.

\section{Examples of Synthetic NCLEX MCQ}
\label{appendix:nclex_examples}

In Table~\ref{table:nclex_examples}, we present one example of a synthetic NCLEX MCQ per specialization generated by GPT-4o.

\begin{table*}[!h]
\centering
\resizebox{\linewidth}{!}{
\begin{tabular}{llllllll}
\hline
\textbf{Question}                                                                                                                                                                                                                                                                                                                                                                                   & \textbf{Choices}                                                                                                                                                                                                                                                                                                            & \textbf{Answer} & \textbf{Rationale}                                                                                                                                                                                                                                                                                                                                                                                                                                                                                         & \textbf{Specialization}                                                             & \textbf{Domain}                                                                              & \textbf{Topic}                                                                & \textbf{Concept}                                                                                 \\ \hline
\begin{tabular}[c]{@{}l@{}}A nurse is assessing a postpartum client\\ who delivered 3 days ago.\\ Which type of lochia should the nurse\\ expect to find at this stage?\end{tabular}                                                                                                                                                                                                                & \begin{tabular}[c]{@{}l@{}}"Lochia rubra", \\ "Lochia serosa", \\ "Lochia alba", \\ "Lochia sanguineous"\end{tabular}                                                                                                                                                                                                       & 0               & \begin{tabular}[c]{@{}l@{}}Lochia rubra is the normal discharge observed\\ for the first 3-4 days postpartum. It is bright red\\ in color and contains blood, decidual and\\ trophoblastic debris. Lochia serosa follows after\\ lochia rubra and lasts for 4-10 days. Lochia alba\\ is the final stage, typically occurring from day 10\\ to up to 6 weeks postpartum. There is no type\\ called lochia sanguineous.\end{tabular}                                                                         & \begin{tabular}[c]{@{}l@{}}Special Domains of\\ Nursing\end{tabular}                & \begin{tabular}[c]{@{}l@{}}Obstetric and\\ Neonatal Care\end{tabular}                        & Postpartum                                                                    & lochia types                                                                                     \\
\begin{tabular}[c]{@{}l@{}}A 65-year-old man with a family history\\ of prostate cancer questions the timing of\\ his prostate cancer screening. What factor\\ is most important in determining the\\ appropriate time for screening initiation?\end{tabular}                                                                                                                                       & \begin{tabular}[c]{@{}l@{}}"Age alone",\\ "Family history of prostate cancer",\\ "Presence of urinary symptoms",\\ "Diet and lifestyle"\end{tabular}                                                                                                                                                                        & 1               & \begin{tabular}[c]{@{}l@{}}For men with a family history of prostate cancer,\\ screening can start earlier than the general\\ recommended age of 50. The USPSTF suggests\\ individualized decision-making about screening\\ for men aged 55-69, but those with a strong family\\ history may benefit from starting earlier.\end{tabular}                                                                                                                                                                   & \begin{tabular}[c]{@{}l@{}}Health Promotion\\ and Disease\\ Prevention\end{tabular} & \begin{tabular}[c]{@{}l@{}}Screening and \\ Preventive Health\\ Checks\end{tabular}          & Routine Screenings                                                            & \begin{tabular}[c]{@{}l@{}}Cancer screenings\\ (e.g., mammograms, \\ colonoscopies)\end{tabular} \\
\begin{tabular}[c]{@{}l@{}}A patient diagnosed with terminal cancer\\ expresses the desire to refuse further\\ chemotherapy. Which document should\\ the nurse refer to in order to honor the\\ patient's wishes?\end{tabular}                                                                                                                                                                      & \begin{tabular}[c]{@{}l@{}}"Medical history", \\ "Advance directive", \\ "Insurance policy", \\ "Patient education booklet"\end{tabular}                                                                                                                                                                                    & 1               & \begin{tabular}[c]{@{}l@{}}document that allows patients to specify their\\ preferences for medical treatment if they\\ become unable to communicate their decisions. In\\ this case, the advance directive would include the\\ patient's wish to refuse further chemotherapy, guiding\\ healthcare providers in respecting the patient's\\ autonomy and decisions regarding their treatment.\end{tabular}                                                                                                 & \begin{tabular}[c]{@{}l@{}}Communication and\\ Interpersonal Skills\end{tabular}    & Ethics and Legal                                                                             & \begin{tabular}[c]{@{}l@{}}Ethical and \\ Legal Issues\end{tabular}           & advance directive                                                                                \\
\begin{tabular}[c]{@{}l@{}}A nurse is evaluating the effectiveness\\ of pain management in a patient who\\ underwent abdominal surgery. The\\ patient reports a pain level of 2 on a\\ scale of 0-10, has a relaxed facial\\ expression, and is able to participate\\ in physical therapy exercises. Which\\ it the most appropriate evaluation of\\ pain management for this patient?\end{tabular} & \begin{tabular}[c]{@{}l@{}}"Pain management is partially effective; further evaluation is needed.",\\ "Pain management is ineffective; the pain level is still present.",\\ "Pain management is effective; pain is well-controlled.", \\ "Pain management needs to be revised entirely due to continued pain."\end{tabular} & 2               & \begin{tabular}[c]{@{}l@{}}Evaluation of pain management involves assessing\\ both subjective and objective indicators of pain relief.\\ Reporting a low pain level (2/10), having a relaxed\\ facial expression, and participating in physical\\ activities are all indicative of well-managed pain.\\ Hence, it is reasonable to conclude that pain\\ management is effective.\end{tabular}                                                                                                              & \begin{tabular}[c]{@{}l@{}}Professional Nursing\\ Practice\end{tabular}             & \begin{tabular}[c]{@{}l@{}}Evidence-Based\\ Practice and\\ Research\end{tabular}             & \begin{tabular}[c]{@{}l@{}}Clinical Practice\\ Guidelines\end{tabular}        & evaluation                                                                                       \\
\begin{tabular}[c]{@{}l@{}}During a health assessment, a nurse\\ notes an abnormal heart rhythm.\\ Which action should the nurse\\ document?\end{tabular}                                                                                                                                                                                                                                           & \begin{tabular}[c]{@{}l@{}}"The time the abnormal rhythm was first noted.", \\ "A probable diagnosis based on the abnormal rhythm.", \\ "How the patient feels about the abnormal rhythm.", \\ "The abnormal rhythm and immediately notify the family."\end{tabular}                                                        & 0               & \begin{tabular}[c]{@{}l@{}}Accurate documentation of the exact time\\ the abnormal rhythm was first detected is crucial\\ for medical treatment decisions and continuity of\\ care. It helps in tracking the condition and correlating\\ it with other clinical signs and symptoms.\end{tabular}                                                                                                                                                                                                           & \begin{tabular}[c]{@{}l@{}}Clinical Skills\\ and Knowledge\end{tabular}             & Patient Assessment                                                                           & \begin{tabular}[c]{@{}l@{}}Health and Physical\\ Assessment\end{tabular}      & \begin{tabular}[c]{@{}l@{}}documenting health\\ assessment\end{tabular}                          \\
\begin{tabular}[c]{@{}l@{}}A nurse working in an urban clinic\\ notices that many patients from \\ low-income neighborhoods have higher\\ rates of chronic diseases. What is the\\ most likely reason for this observation?\end{tabular}                                                                                                                                                            & \begin{tabular}[c]{@{}l@{}}"Lack of genetic variation in the population", \\ "Limited access to healthy food options", \\ "Decreased prevalence of exercise routines", \\ "Inadequate health education programs"\end{tabular}                                                                                               & 1               & \begin{tabular}[c]{@{}l@{}}People living in poverty often reside in areas known\\ as 'food deserts,' where access to affordable, healthy\\ food options is limited. This lack of access can lead\\ to poor nutrition, which is a significant risk factor for\\ chronic diseases such as diabetes, hypertension, and\\ obesity. Addressing this issue involves improving\\ food availability and affordability.\end{tabular}                                                                                & \begin{tabular}[c]{@{}l@{}}Global Health \\ and Nursing\end{tabular}                & \begin{tabular}[c]{@{}l@{}}Health Equity and \\ Social Determinants\\ of Health\end{tabular} & \begin{tabular}[c]{@{}l@{}}Social Determinants\\ of Health (SDH)\end{tabular} & poverty                                                                                          \\
\begin{tabular}[c]{@{}l@{}}A nurse is educating a patient about the\\ benefits of maintaining their own\\ personal health records (PHRs). Which\\ of the following statements made by the\\ patient indicates a need for further\\ teaching?\end{tabular}                                                                                                                                           & \begin{tabular}[c]{@{}l@{}}"I will be able to access my medical history easily.",\\ "I won't need to share my PHR with my healthcare provider.",\\ "It can help me track my progress on health goals.",\\ "I can have all my medication records in one place."\end{tabular}                                                 & 1               & \begin{tabular}[c]{@{}l@{}}While PHRs empower patients by giving them control\\ over their personal health information, these records\\ are most beneficial when shared with healthcare\\ providers. This collaboration ensures a comprehensive\\ understanding of the patient's health, facilitating better\\ management and treatment. Therefore, the statement \\ 'I won't need to share my PHR with my healthcare\\ provider' indicates a misunderstanding of the intended\\ use of PHRs.\end{tabular} & \begin{tabular}[c]{@{}l@{}}Technology and\\ Innovation in\\ Nursing\end{tabular}    & \begin{tabular}[c]{@{}l@{}}Digital Health and\\ Wellness Tools\end{tabular}                  & \begin{tabular}[c]{@{}l@{}}Personal Health\\ Records (PHRs)\end{tabular}      & \begin{tabular}[c]{@{}l@{}}patient-maintained\\ records\end{tabular}                             \\ \hline
\end{tabular}
}
\caption{
\label{table:nclex_examples}
Examples of synthetic NCLEX MCQ generated by GPT-4o.
}
\end{table*}

\section{Example of Reasoning-incorporated Response}
\label{appendix:reasoning_example}

\begin{figure}[!ht]
    \centering
  \includegraphics[width=\columnwidth]{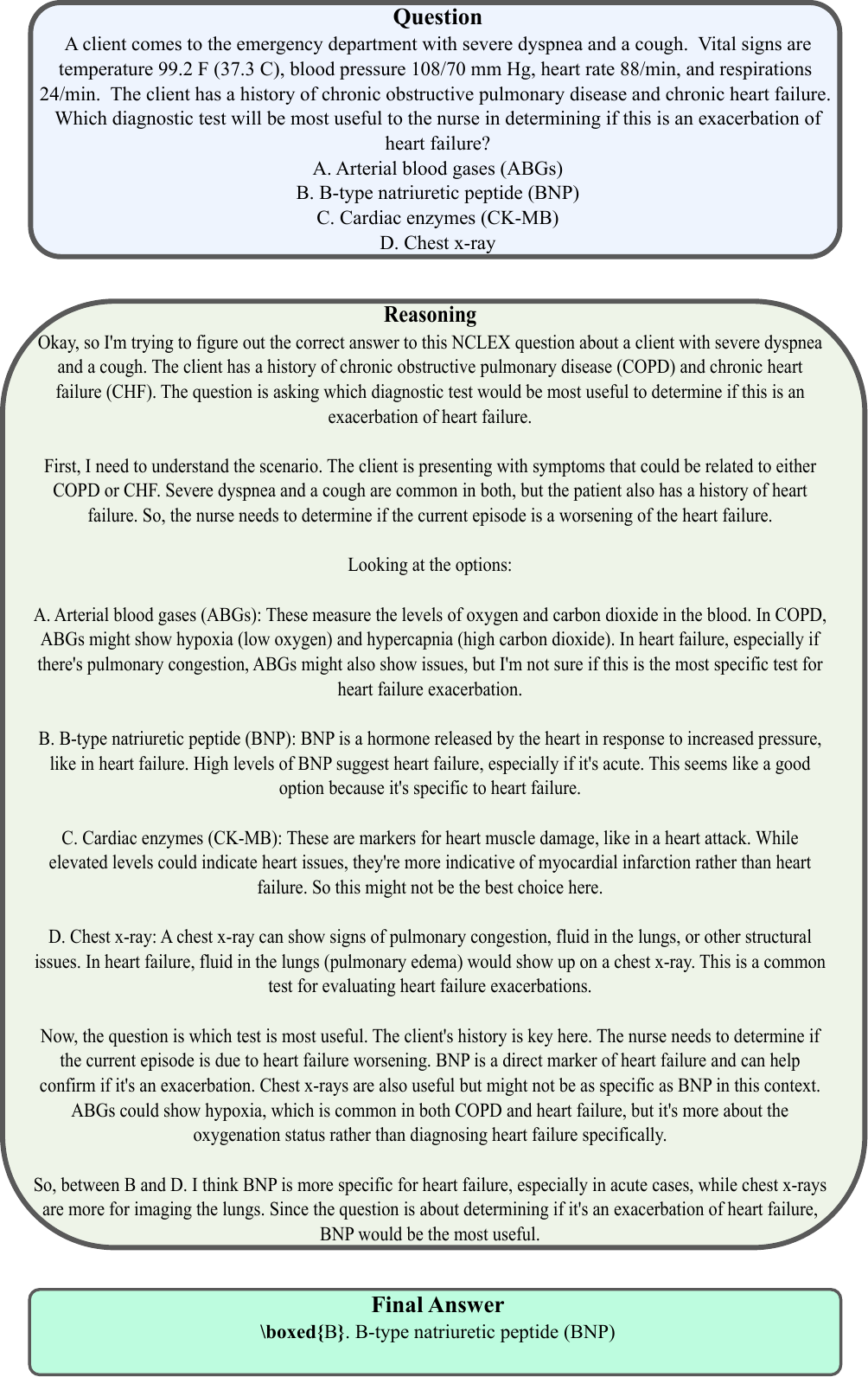}
  \caption{\label{fig:reasoning_example}
  Example of a \nursegpt~response along with the reasoning.
  }
\end{figure}

We present an example of response from \nursegpt~trained with reasoning in Figure~\ref{fig:reasoning_example}.

\section{Sample Response from Multi-Agent System}
\label{appendix:sample_mas}

\begin{table*}[h]
\resizebox{\textwidth}{!}{
\begin{tabular}{l|l}
\hline
\textbf{Question}                                                             & \begin{tabular}[c]{@{}l@{}}During the initial assessment of a client who has sustained a chemical burn to the eye, what additional information is most important for the nurse to obtain?\\ Options:\\ A: The client's pain level on a scale of 1-10.,\\ B: The type and exact composition of the chemical.,\\ C: Whether the client has a history of eye problems.,\\ D:The time elapsed since the exposure.\end{tabular}                                                                                                                                                                                                                                                                                                                                                                                                                                                                                                                                                                                                                                                                                                                                                                                                                                                                                                                                                                                                                                                                                                                                                                                                                                                                                                                                                                                                                                                                                                                                                                                                                                                                                                                                                                                                                                                                                                                                                                                                                                                                                                                                                                                                                                                                                                                                                                                                                                                                                                                                                                                                                                                                                                                                                                                                                                                                                                                                                                                                                                                                                                                                \\ \hline
\textbf{Orchestrator}                                                         & \begin{tabular}[c]{@{}l@{}}Recruitment: Agent 1 (1. Emergency Room Nurse): Specializes in stabilizing trauma patients and managing acute conditions, including chemical exposures. \\ Agent 2 ( 2. Ophthalmic Nurse): Focuses on the care of patients with eye conditions and injuries. \\ Agent 3 ( 3. Toxicology Nurse): Specializes in managing patients with toxic exposures and understanding the effects of various chemicals. \\ Agent 4 ( 4. Critical Care Nurse (ICU Nurse)): Specializes in caring for critically ill patients who may require advanced interventions. \\ Agent 5 ( 5. Pain Management Nurse): Focuses on assessing and managing pain levels in patients, including those with acute injuries.\end{tabular}                                                                                                                                                                                                                                                                                                                                                                                                                                                                                                                                                                                                                                                                                                                                                                                                                                                                                                                                                                                                                                                                                                                                                                                                                                                                                                                                                                                                                                                                                                                                                                                                                                                                                                                                                                                                                                                                                                                                                                                                                                                                                                                                                                                                                                                                                                                                                                                                                                                                                                                                                                                                                                                                                                                                                                                                                     \\ \hline
\textbf{\begin{tabular}[c]{@{}l@{}}Initial\\ Report\end{tabular}}             & \begin{tabular}[c]{@{}l@{}}(emergency room nurse): Answer: (B) The type and exact composition of the chemical. Reason: Knowing the specific type and composition of the chemical\\ that caused the burn is crucial for determining the appropriate treatment and possible interventions. Different chemicals can cause varying degrees of damage\\ and may require specific antidotes or treatments. This information will guide the nurse and the healthcare team in managing the chemical burn effectively and\\ minimizing potential complications. \\ (ophthalmic nurse): Answer: (B) The type and exact composition of the chemical. Reason: Obtaining information about the type and exact composition of the\\ chemical that caused the burn is critical in order to determine the appropriate treatment and interventions. Different chemicals can have varying effects on the\\ eye and may require specific antidotes or management strategies to mitigate damage and preserve vision. This information is essential for ensuring proper and\\ timely medical care. \\ (toxicology nurse): Answer: (B) The type and exact composition of the chemical. Reason: Knowing the type and exact composition of the chemical involved\\ in the eye burn is crucial for determining the appropriate treatment and interventions. Different chemicals can cause varying degrees of damage and require\\ specific decontamination and management strategies. This information will guide the nurse and the healthcare team in providing the most effective care for\\ the client. \\ (critical care nurse (icu nurse)): Answer: (B) The type and exact composition of the chemical. Reason: Determining the type and exact composition of the\\ chemical involved in the eye burn is crucial for guiding the appropriate treatment plan. Different chemicals can have varying effects on the eye and may\\ require specific antidotes or management strategies. This information will directly impact the urgency and focus of interventions, ensuring the best possible\\ outcome for the client. \\ (pain management nurse): Answer: (B) The type and exact composition of the chemical. Reason: Obtaining information about the type and exact\\ composition of the chemical is crucial for determining the appropriate treatment and potential complications related to eye exposure. Different chemicals\\ can have varying effects on ocular tissues, and knowing this information helps guide immediate interventions and management strategies to mitigate damage\\ and promote healing.\end{tabular}                                                                                                                                                                                                                                                                                                                                                                                                                                                                                                                                                                                                                                                                                                                                                                                                                                                                                                                                             \\ \hline
\textbf{\begin{tabular}[c]{@{}l@{}}Partial\\ Conversation\\ Log\end{tabular}} & \begin{tabular}[c]{@{}l@{}}Agent 1 -\textgreater Agent 3: As an emergency room nurse specializing in stabilizing trauma patients and managing acute conditions, including chemical exposures,\\ I want to emphasize the importance of knowing the type and exact composition of the chemical involved in the eye burn. To the toxicology nurse: Given your\\ expertise in managing patients with toxic exposures, I believe it’s crucial for us to work closely together to determine the chemical's identity. This information\\ will significantly influence our treatment strategies, including the selection of specific antidotes or decontamination protocols. Different chemicals can have\\ varying effects on ocular and systemic health, so understanding the composition will be vital for optimizing patient care and minimizing potential complications.\\ Your insight in this area could guide our immediate interventions effectively.\\ Agent 2 -\textgreater Agent 3: As an ophthalmic nurse focusing on the care of patients with eye conditions and injuries, I would like to address the toxicology nurse.\\ Opinion to Agent 3 (Toxicology Nurse): Given the critical role of understanding the specific chemical exposure in managing a chemical burn to the eye, I\\ emphasize the importance of obtaining the type and exact composition of the chemical involved. This information is pivotal not only for determining the\\ appropriate treatment regimen but also for selecting any necessary antidotes or decontamination procedures. The varying effects of different chemicals on ocular\\ tissues can lead to significant differences in treatment protocols. Therefore, your expertise in toxicology will be invaluable in guiding our approach to effectively\\ mitigate the risks and complications associated with this eye injury.\\ Agent 3  -\textgreater Agent 2: As a toxicology nurse specializing in managing patients with toxic exposures, I would like to share my opinion with Agent 2, the ophthalmic\\ nurse. Opinion: It is essential to obtain the type and exact composition of the chemical that caused the eye burn. Reason: Different chemicals have unique\\ mechanisms of injury and may require specific treatment protocols. Knowing the exact chemical will guide us in administering the right decontamination and\\ therapeutic interventions to minimize ocular damage and preserve the patient's vision. Timely and accurate treatment, based on this critical information, is vital\\ for optimal outcomes in chemical eye injuries.Agent 5 -\textgreater Agent 3: As a pain management nurse, my specialty lies in assessing and managing pain levels in\\ patients, including those with acute injuries. To the toxicology nurse: I believe it's imperative that we focus on understanding the type and exact composition of\\ the chemical involved in this eye burn. This information is not only crucial for tailoring urgent treatment protocols but also for anticipating potential complications\\ related to specific toxic substances. Different chemicals may require different management strategies, and knowing the chemical composition will greatly enhance\\ our ability to administer appropriate antidotes or decontamination measures. This can significantly impact the outcome for the patient, including pain management\\ and the preservation of vision. Your expertise in toxicology will be invaluable in guiding our interventions effectively.\end{tabular} \\ \hline
\textbf{\begin{tabular}[c]{@{}l@{}}Moderator\\ Decision\end{tabular}}         & \begin{tabular}[c]{@{}l@{}}Based on the opinions provided by the nursing specialists, the majority consensus is as follows:\\ **Emergency Room Nurse**: (B) The type and exact composition of the chemical.\\ **Ophthalmic Nurse**: (B) The type and exact composition of the chemical.\\ **Toxicology Nurse**: (B) The type and exact composition of the chemical.\\ **Critical Care Nurse (ICU Nurse)**: (B) The type and exact composition of the chemical.\\ **Pain Management Nurse**: (B) The type and exact composition of the chemical.\\ All five specialists unanimously agree that the most important additional information to obtain during the initial assessment of a client with a chemical burn to the\\ eye is the type and exact composition of the chemical.\\ Final Decision:\\ Answer: (B) The type and exact composition of the chemical.\end{tabular}                                                                                                                                                                                                                                                                                                                                                                                                                                                                                                                                                                                                                                                                                                                                                                                                                                                                                                                                                                                                                                                                                                                                                                                                                                                                                                                                                                                                                                                                                                                                                                                                                                                                                                                                                                                                                                                                                                                                                                                                                                                                                                                                                                                                                                                                                                                                                                                                                                                                                                                                                                                                                                                                             \\ \hline
\end{tabular}

}
\caption{
\label{table:sample_mas}
Response from the multi-agent system (due to the space constraint, we only show partial conversation log).
}
\end{table*}

We show sample response from our multi-agent system in Table~\ref{table:sample_mas}.

\clearpage
\newpage
\section{List of Topics used for Data Generation}
\label{appendix:topic_list}

We present the complete list of topic that we create to generate diverse synthetic NCLEX samples below.

\begin{table*}[t]
\centering
\resizebox{\linewidth}{!}{
\begin{tabular}{llll}
\hline
\multicolumn{1}{l}{\textbf{Specialization}} & \multicolumn{1}{l}{\textbf{Domain}}              & \multicolumn{1}{l}{\textbf{Topic}}                        & \multicolumn{1}{l}{\textbf{Concept}}\\
\hline
Clinical Skills and Knowledge               & Patient Assessment                               & Vital Signs Monitoring                                     & Temperature assessment                                                                                           \\
Clinical Skills and Knowledge               & Patient Assessment                               & Vital Signs Monitoring                                     & pulse assessment                                                                                                 \\
Clinical Skills and Knowledge               & Patient Assessment                               & Vital Signs Monitoring                                     & pulse deficit                                                                                                    \\
Clinical Skills and Knowledge               & Patient Assessment                               & Vital Signs Monitoring                                     & respiration assessment                                                                                           \\
Clinical Skills and Knowledge               & Patient Assessment                               & Vital Signs Monitoring                                     & Calculations for Pulse pressure + MAP + CPP                                                                      \\
Clinical Skills and Knowledge               & Patient Assessment                               & Vital Signs Monitoring                                     & blood pressure assessment                                                                                        \\
\hline
\end{tabular}
}
\caption{
\label{table:topic_list_13_2}
Topic list used for synthetic data generation.
}
\end{table*}

\begin{table*}[t]
\centering
\resizebox{\linewidth}{!}{
\begin{tabular}{llll}
\hline
\multicolumn{1}{l}{\textbf{Specialization}} & \multicolumn{1}{l}{\textbf{Domain}}              & \multicolumn{1}{l}{\textbf{Topic}}                        & \multicolumn{1}{l}{\textbf{Concept}}\\
\hline

Clinical Skills and Knowledge               & Patient Assessment                               & Health History Taking                                      & Munchausen (by proxy) syndrome and how to handle that                                                            \\
Clinical Skills and Knowledge               & Patient Assessment                               & Health History Taking                                      & Techniques for documentation.                                                                                    \\
Clinical Skills and Knowledge               & Medication Administration                        & Dosage Calculations                                        & common dosage calculations                                                                                       \\
Clinical Skills and Knowledge               & Medication Administration                        & Dosage Calculations                                        & Mathematical skills for accurate dosage calculations.                                                            \\
Clinical Skills and Knowledge               & Medication Administration                        & Complex Pharmacological Concepts                           & management of sedation, agitation, and delirium                                                                  \\
Clinical Skills and Knowledge               & Medication Administration                        & Complex Pharmacological Concepts                           & richmond agitation sedation scale (RAAS)                                                                         \\
\hline
\end{tabular}
}
\caption{
\label{table:topic_list_13_3}
Topic list used for synthetic data generation.
}
\end{table*}

\begin{table*}[t]
\centering
\resizebox{\linewidth}{!}{
\begin{tabular}{llll}
\hline
\multicolumn{1}{l}{\textbf{Specialization}} & \multicolumn{1}{l}{\textbf{Domain}}              & \multicolumn{1}{l}{\textbf{Topic}}                        & \multicolumn{1}{l}{\textbf{Concept}}\\
\hline

Clinical Skills and Knowledge               & Medication Administration                        & Complex Care Concepts                                      & gastrostomy tube                                                                                                 \\
Clinical Skills and Knowledge               & Medication Administration                        & Complex Care Concepts                                      & jejunostomy tube                                                                                                 \\
Clinical Skills and Knowledge               & Medication Administration                        & Complex Care Concepts                                      & intestinal tubes                                                                                                 \\
Clinical Skills and Knowledge               & Medication Administration                        & Complex Care Concepts                                      & esophageal and gastric tubes                                                                                     \\
Clinical Skills and Knowledge               & Medication Administration                        & Complex Care Concepts                                      & lavage tubes                                                                                                     \\
Clinical Skills and Knowledge               & Medication Administration                        & Complex Care Concepts                                      & urinary and renal tubes                                                                                          \\
\hline
\end{tabular}
}
\caption{
\label{table:topic_list_13_4}
Topic list used for synthetic data generation.
}
\end{table*}

\begin{table*}[t]
\centering
\resizebox{\linewidth}{!}{
\begin{tabular}{llll}
\hline
\multicolumn{1}{l}{\textbf{Specialization}} & \multicolumn{1}{l}{\textbf{Domain}}              & \multicolumn{1}{l}{\textbf{Topic}}                        & \multicolumn{1}{l}{\textbf{Concept}}\\
\hline

Clinical Skills and Knowledge               & Lab Test                                         & Laboratory Values                                          & hemoglobin and hematocrit                                                                                        \\
Clinical Skills and Knowledge               & Lab Test                                         & Laboratory Values                                          & lipids                                                                                                           \\
Clinical Skills and Knowledge               & Lab Test                                         & Laboratory Values                                          & fasting blood glucose                                                                                            \\
Clinical Skills and Knowledge               & Lab Test                                         & Laboratory Values                                          & HgbA1C                                                                                                           \\
Clinical Skills and Knowledge               & Lab Test                                         & Laboratory Values                                          & renal function studies                                                                                           \\
Clinical Skills and Knowledge               & Lab Test                                         & Laboratory Values                                          & blood urea nitrogen (BUN)                                                                                        \\
\hline
\end{tabular}
}
\caption{
\label{table:topic_list_13_5}
Topic list used for synthetic data generation.
}
\end{table*}

\begin{table*}[t]
\centering
\resizebox{\linewidth}{!}{
\begin{tabular}{llll}
\hline
\multicolumn{1}{l}{\textbf{Specialization}} & \multicolumn{1}{l}{\textbf{Domain}}              & \multicolumn{1}{l}{\textbf{Topic}}                        & \multicolumn{1}{l}{\textbf{Concept}}\\
\hline

Clinical Skills and Knowledge               & Integumentary Nursing                            & Integumentary Problems of the Adult Client                 & melanoma                                                                                                         \\
Clinical Skills and Knowledge               & Integumentary Nursing                            & Integumentary Problems of the Adult Client                 & squamous cell carcinoma                                                                                          \\
Clinical Skills and Knowledge               & Integumentary Nursing                            & Integumentary Problems of the Adult Client                 & basal cell carcinoma                                                                                             \\
Clinical Skills and Knowledge               & Integumentary Nursing                            & Integumentary Medications                                  & poison ivy treatment                                                                                             \\
Clinical Skills and Knowledge               & Integumentary Nursing                            & Integumentary Medications                                  & dermatitis treatment                                                                                             \\
Clinical Skills and Knowledge               & Integumentary Nursing                            & Integumentary Medications                                  & topical glucocorticoids                                                                                          \\
\hline
\end{tabular}
}
\caption{
\label{table:topic_list_13_6}
Topic list used for synthetic data generation.
}
\end{table*}

\begin{table*}[t]
\centering
\resizebox{\linewidth}{!}{
\begin{tabular}{llll}
\hline
\multicolumn{1}{l}{\textbf{Specialization}} & \multicolumn{1}{l}{\textbf{Domain}}              & \multicolumn{1}{l}{\textbf{Topic}}                        & \multicolumn{1}{l}{\textbf{Concept}}\\
\hline

Clinical Skills and Knowledge               & Mental Health Nursing                            & Mental Health Problems of the Adult Client                 & family therapy                                                                                                   \\
Clinical Skills and Knowledge               & Mental Health Nursing                            & Mental Health Problems of the Adult Client                 & therapeutic and nontherapeutic communication techniques                                                          \\
Clinical Skills and Knowledge               & Mental Health Nursing                            & Mental Health Problems of the Adult Client                 & types of defence mechanisms                                                                                      \\
Clinical Skills and Knowledge               & Mental Health Nursing                            & Mental Health Problems of the Adult Client                 & client rights                                                                                                    \\
Clinical Skills and Knowledge               & Mental Health Nursing                            & Mental Health Problems of the Adult Client                 & self-help and support groups                                                                                     \\
Clinical Skills and Knowledge               & Mental Health Nursing                            & Mental Health Problems                                     & anxiety                                                                                                          \\
\hline
\end{tabular}
}
\caption{
\label{table:topic_list_13_7}
Topic list used for synthetic data generation.
}
\end{table*}

\begin{table*}[t]
\centering
\resizebox{\linewidth}{!}{
\begin{tabular}{llll}
\hline
\multicolumn{1}{l}{\textbf{Specialization}} & \multicolumn{1}{l}{\textbf{Domain}}              & \multicolumn{1}{l}{\textbf{Topic}}                        & \multicolumn{1}{l}{\textbf{Concept}}\\
\hline

Clinical Skills and Knowledge               & Mental Health Nursing                            & Crisis Theory and Intervention                             & grief                                                                                                            \\
Clinical Skills and Knowledge               & Mental Health Nursing                            & Crisis Theory and Intervention                             & loss                                                                                                             \\
Clinical Skills and Knowledge               & Mental Health Nursing                            & Crisis Theory and Intervention                             & nurse's role in grief and loss                                                                                   \\
Clinical Skills and Knowledge               & Mental Health Nursing                            & Crisis Theory and Intervention                             & suicidal behaviour                                                                                               \\
Clinical Skills and Knowledge               & Mental Health Nursing                            & Crisis Theory and Intervention                             & cues                                                                                                             \\
Clinical Skills and Knowledge               & Mental Health Nursing                            & Crisis Theory and Intervention                             & assessment of suicidal ideation                                                                                  \\
\hline
\end{tabular}
}
\caption{
\label{table:topic_list_13_8}
Topic list used for synthetic data generation.
}
\end{table*}

\begin{table*}[t]
\centering
\resizebox{\linewidth}{!}{
\begin{tabular}{llll}
\hline
\multicolumn{1}{l}{\textbf{Specialization}} & \multicolumn{1}{l}{\textbf{Domain}}              & \multicolumn{1}{l}{\textbf{Topic}}                        & \multicolumn{1}{l}{\textbf{Concept}}\\
\hline

Clinical Skills and Knowledge               & Oncology Nursing                                 & Oncological and Hematological Problems of the Adult Client & risk factors                                                                                                     \\
Clinical Skills and Knowledge               & Oncology Nursing                                 & Oncological and Hematological Problems of the Adult Client & prevention                                                                                                       \\
Clinical Skills and Knowledge               & Oncology Nursing                                 & Oncological and Hematological Problems of the Adult Client & early detection                                                                                                  \\
Clinical Skills and Knowledge               & Oncology Nursing                                 & Oncological and Hematological Problems of the Adult Client & diagnostics                                                                                                      \\
Clinical Skills and Knowledge               & Oncology Nursing                                 & Oncological and Hematological Problems of the Adult Client & pain control                                                                                                     \\
Clinical Skills and Knowledge               & Oncology Nursing                                 & Oncological and Hematological Problems of the Adult Client & surgery                                                                                                          \\
\hline
\end{tabular}
}
\caption{
\label{table:topic_list_13_9}
Topic list used for synthetic data generation.
}
\end{table*}

\begin{table*}[t]
\centering
\resizebox{\linewidth}{!}{
\begin{tabular}{llll}
\hline
\multicolumn{1}{l}{\textbf{Specialization}} & \multicolumn{1}{l}{\textbf{Domain}}              & \multicolumn{1}{l}{\textbf{Topic}}                        & \multicolumn{1}{l}{\textbf{Concept}}\\
\hline

Clinical Skills and Knowledge               & Surgical Nursing                                 & Perioperative Nursing Care                                 & preoperative checklist                                                                                           \\
Clinical Skills and Knowledge               & Surgical Nursing                                 & Perioperative Nursing Care                                 & postoperative care                                                                                               \\
Clinical Skills and Knowledge               & Surgical Nursing                                 & Perioperative Nursing Care                                 & risks of surgery                                                                                                 \\
Clinical Skills and Knowledge               & Surgical Nursing                                 & Perioperative Nursing Care                                 & management of surgical patients                                                                                  \\
Clinical Skills and Knowledge               & Surgical Nursing                                 & Perioperative Nursing Care                                 & wound dehiscence and evisceration                                                                                \\
Clinical Skills and Knowledge               & Surgical Nursing                                 & Perioperative Nursing Care                                 & deep breathing and coughing                                                                                      \\
\hline
\end{tabular}
}
\caption{
\label{table:topic_list_13_10}
Topic list used for synthetic data generation.
}
\end{table*}

\begin{table*}[t]
\centering
\resizebox{\linewidth}{!}{
\begin{tabular}{llll}
\hline
\multicolumn{1}{l}{\textbf{Specialization}} & \multicolumn{1}{l}{\textbf{Domain}}              & \multicolumn{1}{l}{\textbf{Topic}}                        & \multicolumn{1}{l}{\textbf{Concept}}\\
\hline

Professional Nursing Practice               & Interprofessional Collaboration                  & Principles of Interprofessional Practice                   & challenges                                                                                                       \\
Professional Nursing Practice               & Interprofessional Collaboration                  & Principles of Interprofessional Practice                   & strategies for effective interprofessional practice                                                              \\
Professional Nursing Practice               & Interprofessional Collaboration                  & Roles and Responsibilities                                 & Understanding and respecting the roles of different healthcare professionals                                     \\
Professional Nursing Practice               & Interprofessional Collaboration                  & Collaborative Patient-Centered Care                        & Approaches to ensure care is coordinated, comprehensive, and patient-centered                                    \\
Professional Nursing Practice               & Client-focused Care                              & Care of Special Populations                                & health care considerations for the LGBTQ population                                                              \\
Professional Nursing Practice               & Client-focused Care                              & Care of Special Populations                                & special populations needs assessment tools                                                                       \\
\hline
\end{tabular}
}
\caption{
\label{table:topic_list_13_11}
Topic list used for synthetic data generation.
}
\end{table*}

\begin{table*}[t]
\centering
\resizebox{\linewidth}{!}{
\begin{tabular}{llll}
\hline
\multicolumn{1}{l}{\textbf{Specialization}} & \multicolumn{1}{l}{\textbf{Domain}}              & \multicolumn{1}{l}{\textbf{Topic}}                        & \multicolumn{1}{l}{\textbf{Concept}}\\
\hline

Special Domains of Nursing                  & Cardiac Nursing                                  & Interventions                                              & patient education on lifestyle modifications                                                                     \\
Special Domains of Nursing                  & Cardiovascular Nursing                           & cardiovascular problems                                    & diagnostic tests and procedures                                                                                  \\
Special Domains of Nursing                  & Cardiovascular Nursing                           & cardiovascular problems                                    & normal versus abnormal heart sounds                                                                              \\
Special Domains of Nursing                  & Cardiovascular Nursing                           & cardiovascular problems                                    & coronary artery disease                                                                                          \\
Special Domains of Nursing                  & Cardiovascular Nursing                           & cardiovascular problems                                    & angina                                                                                                           \\
Special Domains of Nursing                  & Cardiovascular Nursing                           & cardiovascular problems                                    & myocardial infarction                                                                                            \\
\hline
\end{tabular}
}
\caption{
\label{table:topic_list_13_12}
Topic list used for synthetic data generation.
}
\end{table*}

\begin{table*}[t]
\centering
\resizebox{\linewidth}{!}{
\begin{tabular}{llll}
\hline
\multicolumn{1}{l}{\textbf{Specialization}} & \multicolumn{1}{l}{\textbf{Domain}}              & \multicolumn{1}{l}{\textbf{Topic}}                        & \multicolumn{1}{l}{\textbf{Concept}}\\
\hline

Special Domains of Nursing                  & Cardiovascular Nursing                           & therapeutic management of cardiovascular problems          & atherectomy                                                                                                      \\
Special Domains of Nursing                  & Cardiovascular Nursing                           & therapeutic management of cardiovascular problems          & transmyocardial revascularization                                                                                \\
Special Domains of Nursing                  & Cardiovascular Nursing                           & therapeutic management of cardiovascular problems          & coronary artery bypass grafting                                                                                  \\
Special Domains of Nursing                  & Cardiovascular Nursing                           & therapeutic management of cardiovascular problems          & heart transplantation                                                                                            \\
Special Domains of Nursing                  & Cardiovascular Nursing                           & management of dysrhythmias                                 & vagal maneuvers                                                                                                  \\
Special Domains of Nursing                  & Cardiovascular Nursing                           & management of dysrhythmias                                 & valsalva maneuver                                                                                                \\
\hline
\end{tabular}
}
\caption{
\label{table:topic_list_13_13}
Topic list used for synthetic data generation.
}
\end{table*}

\begin{table*}[t]
\centering
\resizebox{\linewidth}{!}{
\begin{tabular}{llll}
\hline
\multicolumn{1}{l}{\textbf{Specialization}} & \multicolumn{1}{l}{\textbf{Domain}}              & \multicolumn{1}{l}{\textbf{Topic}}                        & \multicolumn{1}{l}{\textbf{Concept}}\\
\hline

Special Domains of Nursing                  & Eye and Ear Nursing                              & Eye and Ear Problems of the Adult Client                   & nursing considerations for retinal detachment                                                                    \\
Special Domains of Nursing                  & Eye and Ear Nursing                              & Eye and Ear Problems of the Adult Client                   & macular degeneration                                                                                             \\
Special Domains of Nursing                  & Eye and Ear Nursing                              & Eye and Ear Problems of the Adult Client                   & ocular melanoma                                                                                                  \\
Special Domains of Nursing                  & Eye and Ear Nursing                              & Eye and Ear Problems of the Adult Client                   & enucleation and exenteration                                                                                     \\
Special Domains of Nursing                  & Eye and Ear Nursing                              & Eye and Ear Problems of the Adult Client                   & hyphema                                                                                                          \\
Special Domains of Nursing                  & Eye and Ear Nursing                              & Eye and Ear Problems of the Adult Client                   & contusions                                                                                                       \\
\hline
\end{tabular}
}
\caption{
\label{table:topic_list_14}
Topic list used for synthetic data generation.
}
\end{table*}

\begin{table*}[t]
\centering
\resizebox{\linewidth}{!}{
\begin{tabular}{llll}
\hline
\multicolumn{1}{l}{\textbf{Specialization}} & \multicolumn{1}{l}{\textbf{Domain}}              & \multicolumn{1}{l}{\textbf{Topic}}                        & \multicolumn{1}{l}{\textbf{Concept}}\\
\hline

Special Domains of Nursing                  & Neurological Nursing                             & Neurological Problems of the Adult Client                  & cerebral aneurysm                                                                                                \\
Special Domains of Nursing                  & Neurological Nursing                             & Neurological Problems of the Adult Client                  & seizure types                                                                                                    \\
Special Domains of Nursing                  & Neurological Nursing                             & Neurological Problems of the Adult Client                  & nursing considerations for a client with seizures                                                                \\
Special Domains of Nursing                  & Neurological Nursing                             & Neurological Problems of the Adult Client                  & stroke                                                                                                           \\
Special Domains of Nursing                  & Neurological Nursing                             & Neurological Problems of the Adult Client                  & types of stroke                                                                                                  \\
Special Domains of Nursing                  & Neurological Nursing                             & Neurological Problems of the Adult Client                  & nursing considerations for a client with stroke                                                                  \\
\hline
\end{tabular}
}
\caption{
\label{table:topic_list_14_2}
Topic list used for synthetic data generation.
}
\end{table*}

\begin{table*}[t]
\centering
\resizebox{\linewidth}{!}{
\begin{tabular}{llll}
\hline
\multicolumn{1}{l}{\textbf{Specialization}} & \multicolumn{1}{l}{\textbf{Domain}}              & \multicolumn{1}{l}{\textbf{Topic}}                        & \multicolumn{1}{l}{\textbf{Concept}}\\
\hline

Special Domains of Nursing                  & Musculoskeletal Nursing                          & Musculoskeletal problems of the adult client               & traction devices                                                                                                 \\
Special Domains of Nursing                  & Musculoskeletal Nursing                          & Musculoskeletal problems of the adult client               & skeletal traction                                                                                                \\
Special Domains of Nursing                  & Musculoskeletal Nursing                          & Musculoskeletal problems of the adult client               & skin traction                                                                                                    \\
Special Domains of Nursing                  & Musculoskeletal Nursing                          & Musculoskeletal problems of the adult client               & Russell's skin (slling) traction                                                                                 \\
Special Domains of Nursing                  & Musculoskeletal Nursing                          & Musculoskeletal problems of the adult client               & pelvic skin traction                                                                                             \\
Special Domains of Nursing                  & Musculoskeletal Nursing                          & Musculoskeletal problems of the adult client               & balanced suspension traction                                                                                     \\
\hline
\end{tabular}
}
\caption{
\label{table:topic_list_14_3}
Topic list used for synthetic data generation.
}
\end{table*}

\begin{table*}[t]
\centering
\resizebox{\linewidth}{!}{
\begin{tabular}{llll}
\hline
\multicolumn{1}{l}{\textbf{Specialization}} & \multicolumn{1}{l}{\textbf{Domain}}              & \multicolumn{1}{l}{\textbf{Topic}}                        & \multicolumn{1}{l}{\textbf{Concept}}\\
\hline

Special Domains of Nursing                  & Immune Nursing                                   & Immune Problems of the Adult Client                        & products that contain latex                                                                                      \\
Special Domains of Nursing                  & Immune Nursing                                   & Immune Problems of the Adult Client                        & interventions for the client with a latex allergy                                                                \\
Special Domains of Nursing                  & Immune Nursing                                   & Immune Problems of the Adult Client                        & assessment and stages of lyme disease                                                                            \\
Special Domains of Nursing                  & Immune Nursing                                   & Immune Problems of the Adult Client                        & tests used to evaluate progression of human immunodeficiency virus (HIV) infection                               \\
Special Domains of Nursing                  & Immune Nursing                                   & Immune Problems of the Adult Client                        & diagnostic criteria for acquired immunodeficiency syndrome (AIDS)                                                \\
Special Domains of Nursing                  & Immune Nursing                                   & Immune medications                                         & HIV and AIDs medications                                                                                         \\
\hline
\end{tabular}
}
\caption{
\label{table:topic_list_14_4}
Topic list used for synthetic data generation.
}
\end{table*}

\begin{table*}[t]
\centering
\resizebox{\linewidth}{!}{
\begin{tabular}{llll}
\hline
\multicolumn{1}{l}{\textbf{Specialization}} & \multicolumn{1}{l}{\textbf{Domain}}              & \multicolumn{1}{l}{\textbf{Topic}}                        & \multicolumn{1}{l}{\textbf{Concept}}\\
\hline

Special Domains of Nursing                  & Endocrine Care                                   & Endocrine Problems of the Adult Client                     & oral hypoglycemic medications                                                                                    \\
Special Domains of Nursing                  & Endocrine Care                                   & Endocrine Problems of the Adult Client                     & complications of insulin therapy                                                                                 \\
Special Domains of Nursing                  & Endocrine Care                                   & Endocrine Problems of the Adult Client                     & dawn phenomenon                                                                                                  \\
Special Domains of Nursing                  & Endocrine Care                                   & Endocrine Problems of the Adult Client                     & somogyi phenomenon                                                                                               \\
Special Domains of Nursing                  & Endocrine Care                                   & Endocrine Problems of the Adult Client                     & insulin administration                                                                                           \\
Special Domains of Nursing                  & Endocrine Care                                   & Endocrine Problems of the Adult Client                     & self-monitoring blood glucose level                                                                              \\
\hline
\end{tabular}
}
\caption{
\label{table:topic_list_14_5}
Topic list used for synthetic data generation.
}
\end{table*}

\begin{table*}[t]
\centering
\resizebox{\linewidth}{!}{
\begin{tabular}{llll}
\hline
\multicolumn{1}{l}{\textbf{Specialization}} & \multicolumn{1}{l}{\textbf{Domain}}              & \multicolumn{1}{l}{\textbf{Topic}}                        & \multicolumn{1}{l}{\textbf{Concept}}\\
\hline

Special Domains of Nursing                  & Gastrointestinal Care                            & Gastrointestinal Problems in Adults                        & signs of bowel perforation                                                                                       \\
Special Domains of Nursing                  & Gastrointestinal Care                            & Gastrointestinal Medications                               & antacids                                                                                                         \\
Special Domains of Nursing                  & Gastrointestinal Care                            & Gastrointestinal Medications                               & gastric protectants                                                                                              \\
Special Domains of Nursing                  & Gastrointestinal Care                            & Gastrointestinal Medications                               & histamine receptor antagonists                                                                                   \\
Special Domains of Nursing                  & Gastrointestinal Care                            & Gastrointestinal Medications                               & proton pump inhibitors                                                                                           \\
Special Domains of Nursing                  & Gastrointestinal Care                            & Gastrointestinal Medications                               & prokinetic agent                                                                                                 \\
\hline
\end{tabular}
}
\caption{
\label{table:topic_list_14_6}
Topic list used for synthetic data generation.
}
\end{table*}

\begin{table*}[t]
\centering
\resizebox{\linewidth}{!}{
\begin{tabular}{llll}
\hline
\multicolumn{1}{l}{\textbf{Specialization}} & \multicolumn{1}{l}{\textbf{Domain}}              & \multicolumn{1}{l}{\textbf{Topic}}                        & \multicolumn{1}{l}{\textbf{Concept}}\\
\hline

Special Domains of Nursing                  & Obstetric and Neonatal Care                      & Risk Conditions Related to Pregnancy                       & types of abortion                                                                                                \\
Special Domains of Nursing                  & Obstetric and Neonatal Care                      & Risk Conditions Related to Pregnancy                       & interventions for abortion                                                                                       \\
Special Domains of Nursing                  & Obstetric and Neonatal Care                      & Risk Conditions Related to Pregnancy                       & diabetes mellitus                                                                                                \\
Special Domains of Nursing                  & Obstetric and Neonatal Care                      & Risk Conditions Related to Pregnancy                       & disseminated intravascular coagulation (DIC)                                                                     \\
Special Domains of Nursing                  & Obstetric and Neonatal Care                      & Risk Conditions Related to Pregnancy                       & ectopic pregnancy                                                                                                \\
Special Domains of Nursing                  & Obstetric and Neonatal Care                      & Risk Conditions Related to Pregnancy                       & fetal death                                                                                                      \\
\hline
\end{tabular}
}
\caption{
\label{table:topic_list_14_7}
Topic list used for synthetic data generation.
}
\end{table*}

\begin{table*}[t]
\centering
\resizebox{\linewidth}{!}{
\begin{tabular}{llll}
\hline
\multicolumn{1}{l}{\textbf{Specialization}} & \multicolumn{1}{l}{\textbf{Domain}}              & \multicolumn{1}{l}{\textbf{Topic}}                        & \multicolumn{1}{l}{\textbf{Concept}}\\
\hline

Special Domains of Nursing                  & Obstetric and Neonatal Care                      & Postpartum                                                 & normal postpartum vital signs                                                                                    \\
Special Domains of Nursing                  & Obstetric and Neonatal Care                      & Postpartum                                                 & signs and symptoms of postpartum depression                                                                      \\
Special Domains of Nursing                  & Obstetric and Neonatal Care                      & Postpartum                                                 & signs and symptoms of postpartum psychosis                                                                       \\
Special Domains of Nursing                  & Obstetric and Neonatal Care                      & Postpartum Complications                                   & cystitis                                                                                                         \\
Special Domains of Nursing                  & Obstetric and Neonatal Care                      & Postpartum Complications                                   & hematoma management in postpartum                                                                                \\
Special Domains of Nursing                  & Obstetric and Neonatal Care                      & Postpartum Complications                                   & uterine atony                                                                                                    \\
\hline
\end{tabular}
}
\caption{
\label{table:topic_list_16_2}
Topic list used for synthetic data generation.
}
\end{table*}

\begin{table*}[t]
\centering
\resizebox{\linewidth}{!}{
\begin{tabular}{llll}
\hline
\multicolumn{1}{l}{\textbf{Specialization}} & \multicolumn{1}{l}{\textbf{Domain}}              & \multicolumn{1}{l}{\textbf{Topic}}                        & \multicolumn{1}{l}{\textbf{Concept}}\\
\hline

Special Domains of Nursing                  & Respiratory Nursing                              & Respiratory Problems in Pediatrics                         & pneumonia types and management                                                                                   \\
Special Domains of Nursing                  & Respiratory Nursing                              & Respiratory Problems in Pediatrics                         & asthma                                                                                                           \\
Special Domains of Nursing                  & Respiratory Nursing                              & Respiratory Problems in Pediatrics                         & cystic fibrosis                                                                                                  \\
Special Domains of Nursing                  & Respiratory Nursing                              & Respiratory Problems in Pediatrics                         & sudden infant death syndrome (SIDS)                                                                              \\
Special Domains of Nursing                  & Respiratory Nursing                              & Respiratory Problems in Pediatrics                         & foreign body aspiration                                                                                          \\
Special Domains of Nursing                  & Respiratory Nursing                              & Respiratory Problems in Pediatrics                         & tuberculosis                                                                                                     \\
\hline
\end{tabular}
}
\caption{
\label{table:topic_list_16}
Topic list used for synthetic data generation.
}
\end{table*}

\begin{table*}[t]
\centering
\resizebox{\linewidth}{!}{
\begin{tabular}{llll}
\hline
\multicolumn{1}{l}{\textbf{Specialization}} & \multicolumn{1}{l}{\textbf{Domain}}              & \multicolumn{1}{l}{\textbf{Topic}}                        & \multicolumn{1}{l}{\textbf{Concept}}\\
\hline

Special Domains of Nursing                  & Public Health Nursing                            & Epidemiology                                               & Outbreak investigation                                                                                           \\
Special Domains of Nursing                  & Public Health Nursing                            & Epidemiology                                               & surveillance                                                                                                     \\
Special Domains of Nursing                  & Public Health Nursing                            & Epidemiology                                               & community health research                                                                                        \\
Special Domains of Nursing                  & Travel Nursing                                   & Adaptability Skills                                        & Adjusting to different clinical environments                                                                     \\
Special Domains of Nursing                  & Travel Nursing                                   & Adaptability Skills                                        & tips for floating to other units                                                                                 \\
Special Domains of Nursing                  & Travel Nursing                                   & Adaptability Skills                                        & flexibility in assignments                                                                                       \\
\hline
\end{tabular}
}
\caption{
\label{table:topic_list_17}
Topic list used for synthetic data generation.
}
\end{table*}

\begin{table*}[t]
\centering
\resizebox{\linewidth}{!}{
%
}
\caption{
\label{table:topic_list_18}
Topic list used for synthetic data generation.
}
\end{table*}



\end{document}